\documentclass{article} % For LaTeX2e
\usepackage{iclr2024_conference,times}

\usepackage{url}            % simple URL typesetting
\usepackage{booktabs}       % professional-quality tables
\usepackage{amsfonts}       % blackboard math symbols
\usepackage{nicefrac}       % compact symbols for 1/2, etc.
\usepackage{microtype}      % microtypography
\usepackage{pifont}
\usepackage{adjustbox}
\usepackage{wrapfig}
\usepackage{multirow,mathtools } 
\usepackage{algorithm,algpseudocode}
\usepackage{rotating}
\usepackage{lscape}
\usepackage{longtable}
\usepackage{subcaption}
\usepackage[pagebackref,breaklinks,colorlinks]{hyperref}
\usepackage[inline]{enumitem}

\usepackage{algorithm,algpseudocode}
\algnewcommand{\algorithmicforeach}{\textbf{for each}}
\algdef{SE}[FOR]{ForEach}{EndForEach}[1]
  {\algorithmicforeach\ #1\ \algorithmicdo}% \ForEach{#1}
  {\algorithmicend\ \algorithmicforeach}% \EndForEach

\usepackage{pifont}
\newcommand{\cmark}{\textcolor{green}{\ding{51}}}
\newcommand{\xmark}{\textcolor{red}{\ding{55}}}

\usepackage{color, colortbl}
\definecolor{Gray}{gray}{0.93}
\definecolor{Orange}{rgb}{1,0.5,0}
\definecolor{DGray}{gray}{0.83}
\definecolor{LightCyan}{rgb}{0.88,1,1}

\definecolor{Sijia_color}{rgb}{0.858, 0.188, 0.478}

\definecolor{Soumyadeep_color}{rgb}{0.3476, 0.6461, 0.3828}

\definecolor{CR}{rgb}{0.7, 0.1, 0.7}

\definecolor{LightCyan}{rgb}{0.78,0.94,1}
\definecolor{Gray}{gray}{0.85}
\definecolor{lightgreen}{rgb}{0.596,0.98,0.596}
\definecolor{verylightgray}{gray}{0.95}
\definecolor{lightred}{rgb}{1.0,0.7,0.7}

\usepackage{wrapfig}

\usepackage{multirow,mathtools } \usepackage{algorithm,algpseudocode}

\usepackage{pifont}
\usepackage{color, colortbl}

\usepackage{blindtext}
\usepackage{lipsum}
\usepackage{soul}

\usepackage{multirow}
\usepackage{graphicx}
\usepackage{listings}

\usepackage{bbm}

\usepackage[most]{tcolorbox}

\usepackage{amsmath,amsfonts,bm}

\def\eqref#1{(\ref{#1})}
\def\1{\bm{1}}

\def\eps{{\epsilon}}

\DeclareMathAlphabet{\mathsfit}{\encodingdefault}{\sfdefault}{m}{sl}
\SetMathAlphabet{\mathsfit}{bold}{\encodingdefault}{\sfdefault}{bx}{n}

\DeclareMathOperator*{\argmax}{arg\,max}
\DeclareMathOperator*{\argmin}{arg\,min}

\DeclareMathOperator*{\ST}{\text{subject to}}

\title{Backdoor Secrets Unveiled:  Identifying Backdoor Data with Optimized Scaled Prediction Consistency
}

\author{Soumyadeep Pal$^1$, Yuguang Yao$^2$, Ren Wang$^3$, Bingquan Shen$^4$, Sijia Liu$^{2,5}$\\
$^1$University of Alberta, $^2$Michigan State University, $^3$Illinois Institute of Technology, \\$^4$DSO National Laboratories, $^5$MIT-IBM Watson AI Lab, IBM Research
}

\iclrfinalcopy % Uncomment for camera-ready version, but NOT for submission.
\begin{document}

\maketitle

\begin{abstract}

Modern machine learning (ML) systems demand substantial training data, often resorting to external sources. Nevertheless, this practice renders them vulnerable to backdoor poisoning attacks. Prior backdoor defense strategies have primarily focused on the identification of backdoored models or poisoned data characteristics, typically operating under the assumption of access to clean data. In this work, we delve into a relatively underexplored challenge: the automatic identification of backdoor data within a poisoned dataset, all under realistic conditions, \textit{i.e.},  without the need for additional clean data or {without} manually defining a threshold for backdoor detection. We draw an inspiration from the {scaled prediction consistency} (SPC) technique, which {exploits}  the prediction invariance of poisoned data to an input scaling factor. Based on this, we {pose}  the backdoor data identification problem as a hierarchical data splitting optimization problem, leveraging a novel SPC-based loss function as the primary optimization objective. Our innovation unfolds in several key aspects. First, we revisit the vanilla SPC method, unveiling its limitations in addressing the proposed backdoor identification problem. Subsequently, we develop a bi-level optimization-based approach to precisely identify backdoor data by minimizing the advanced SPC loss. Finally, we demonstrate the efficacy of our proposal against a spectrum of backdoor attacks, encompassing basic label-corrupted attacks as well as more sophisticated clean-label attacks, evaluated across various benchmark datasets. Experiment results show that our approach often surpasses the performance of current baselines in identifying backdoor data points, resulting in {about 4\%-36\% improvement in average AUROC.} %\st{about an average 4\%-20\% improvement in AUROC}. 
Codes are available at \href{https://github.com/OPTML-Group/BackdoorMSPC}{https://github.com/OPTML-Group/BackdoorMSPC}.

\end{abstract}

\section{Introduction}\label{sec: intro}

% \SL{[General comment, please citep references pricesly and sufficiently. You can use ChatGPT to revise your sentences/paragraph. Yet, please do a proofread and selectly use its revisions.]}

% \SL{[General comment: An overview figure would be good to highlight  results and/or methods.]}

Deep neural networks (DNNs) have been a key component in driving a revolution in artificial intelligence over the past years, with applications in a variety of fields. They are used in computer vision \citep{krizhevsky2017imagenet,goodfellow2020generative,ren2015faster}, autonomous driving \citep{wen2022deep}, face detection \citep{yang2021larnet} and various other realms. However, it has been shown in the literature that such deep networks are often brittle to various kinds of attacks. Attacks on DNNs can be test-time prediction-evasion  attacks  \citep{yuan2019adversarial}, or they can occur through data poisoning attacks \citep{goldblum2022dataset}, which introduce feature and/or label pollution into the training data. In a recent survey conducted among industry professionals \citep{goldblum2022dataset}, it was discovered that the risks associated with data poisoning attacks are continually escalating for practitioners.  In this work, we focus on a specific type of data poisoning attacks, called backdoor attacks. Backdoor attacks are usually performed by an adversary by polluting a small portion of the training dataset with (often) imperceptible backdoor triggers (\textit{e.g.}, a small image patch on  imagery data) such that it establishes a correlation with a target label. When models trained on such data are deployed, they can be easily manipulated by the application of such triggers in the test data.

% \SL{[Modify the paragraph by starting "important avenue of research is developing backdoor defense methods against such attacks." with references, then saying their common limitations, then introducing the problem of "backdoor data sifting or selection".]}

% \SL{[Then, next paragraph stated our motivation for solving this problem and introducing "signatures" of backdoor data, SPC etc. ]}

%  \SL{[Then, you can state the most relevant work to ours, introduce differences with it, and elaborate on our contributions.]}

% With the advent of the deeper and deeper networks, our models need an extensive amount of training data. So practitioners may often scrape data off the internet or acquire data from third party sources. In such a scenario, it is rather easy for an adversary to pollute said training data, thereby performing backdoor attacks. 
Thus, an important avenue of research is developing backdoor defense methods against such attacks. Previous defense methods mostly focus on suppressing the backdoor attack \citep{du2019robust,borgnia2021strong,borgnia2021dp,pal2023towards,hong2020effectiveness,liu2021removing,li2021neural}
% \SL{[more refs if possible]}
or detecting poisoned models \citep{chen2019deepinspect,kolouri2020universal,wang2020practical,shen2021backdoor,xu2021detecting}. % \SL{[more refs if possible]}
% \begin{wrapfigure}{r}{0.7\textwidth}
% \vspace*{-2mm}
% \centerline{
% \includegraphics[width=0.7\textwidth,height=!]{figs/Fig1_RealisticConditions.pdf}
% }
% \caption{\footnotesize{\SP{Figure Not finished.} \textcolor{red}{Comments}}
% }
% \label{fig: real}
% \vspace*{-2.5mm}
% \end{wrapfigure}%%%
% % https://lucid.app/lucidchart/7e450b24-7803-406d-b4a4-427025e0907b/edit?viewport_loc=-536%2C-56%2C2496%2C1266%2C0_0&invitationId=inv_d16646cb-d200-4244-8f11-9197488d27b0
However, a more formidable challenge lies in directly identifying and pinpointing the backdoor samples concealed within the training set. On achieving such identification, users have much more freedom in their practice. In particular, a more intriguing inquiry pertains to how the aforementioned problem can be addressed under \textbf{practical conditions}. These include: \begin{enumerate*}
    \item[\textbf{(P1)}] \textbf{Free of Clean Data}: The user does not have access to \textit{additional} clean base set prior to starting their training.
    \item[\textbf{(P2)}] \textbf{Free of Detection Threshold}: An identification algorithm may assign scores to training data samples which can reveal the likelihood of samples being backdoored, \textit{e.g.}, a score by averaging training loss at early epochs \citep{li2021anti}. We present different scoring methods in Appendix \ref{app: idscore}. In such a scenario, a user would need to manually set a detection threshold for these scores either heuristically or by knowledge of poisoning ratio \citep{zeng2022sift}. We argue that in a realistic scenario, users should not be required to set such a manual threshold. We emphasize the importance of these practical conditions in Appendix \ref{app: prac_imp}.
\end{enumerate*} 
% In \textbf{Figure\,\ref{fig: real}}, we visually depict the focal point of our study: \textit{the identification of backdoor data points under realistic conditions}, highlighting its distinctions from conventional backdoor data identification techniques.
%%%%%%%%%%%%%%%%%%%%%%%%%%%%%%%%%%%%%%%%%%%%%%%%%%%%%%
% \SL{more formidable challenge lies in directly detecting and pinpointing the backdoor samples concealed within the training set, rather than solely identifying the presence of a backdoored model. In particular, 

% \iffalse
% {\textcolor{red}{1. Sifting out / Backdoor Membership 2. Calling it sifting is done in STRIP}}

%\iffalse
% \SL{[see my edits above, and merge the following to the above as well.]}

%%%%%%%%%%%%%%%%%%%%%%%%%%%%%%%%%%%%%%%%%%%%%%%%%%%%%%
% \SL{[consider removing the following Q.]}
% Thus, in this paper, we take a step towards addressing the following challenge: 
% % \SL{[To be honest, detecting/identifying/sifting out backdoor samples are a bit confusing, same meaning in our paper?]}

% \begin{center}
% \vspace*{-1.3mm}
% 	\setlength\fboxrule{0.1pt}
% 	\noindent\fcolorbox{black}[rgb]{0.8,0.8,0.8}{\begin{minipage}{0.96\columnwidth}
% 			%\begin{center}
% 				\vspace{-0.00cm}
% 	\textbf{Q.} Can backdoor data samples be \textit{accurately} identified from a training set in a \textit{realistic setting}?
% 				\vspace{-0.00cm}
% 		%	\end{center}
% 	\end{minipage}}
% 	% \vspace*{mm}
% \end{center}

\begin{wraptable}{r}{0.6\textwidth}
\vspace*{-4mm}
\centering
\caption{\footnotesize{
\textbf{Practical conditions} for different backdoor identification methods. \xmark  \ denotes undesired conditions and \cmark \ denotes desired conditions. 
% Some works were not designed for sifting out backdoor samples, but they can be easily adapted to the sifting scenario. Thus, you can call `Applicability for sifting
} 
}
\label{tab: realtable}
% \resizebox{0.6\textwidth}{!}{
% \begin{tabular}{c|c|c|c}
% \toprule[1pt]
% \midrule
% \multirow{2}{*}{Defense Method} &  \multicolumn{3}{c}{Conditions} \\
% & Sifting  & Without Clean Data &Without Threshold  \\
% \midrule
%  STRIP &  \cmark & \xmark & \xmark \\
%  ABL &  \xmark & \cmark & \cmark \\
%     DBD &  \xmark & \cmark & \cmark \\
%  SPC &  \cmark & - & \xmark \\
%  CD &  \cmark & \xmark & \xmark \\
%  SD-FCT &  \xmark & \cmark & \xmark \\
% \midrule
%  Scale-SIFT (Ours) &  \cmark & \cmark & \cmark \\
% \bottomrule[1pt]
% \end{tabular}
% }
% \vspace*{-4mm}
% \end{wraptable}%%%
\resizebox{0.6\textwidth}{!}{
\begin{tabular}{c|c|c|c}
\toprule[1pt]
\midrule
\multirow{3}{*}{Backdoor Defenses} & Backdoor & \multicolumn{2}{c}{Practical Conditions} \\
& Data & \textbf{(P1)} Free of & \textbf{(P2)} Free of  \\
& Identification & Clean Data &  Detection Threshold  \\
\midrule
ABL \citep{li2021anti} &  \xmark & \cmark & \xmark \\
DBD \citep{huang2022backdoor} &  \xmark & \cmark & \xmark \\
SD-FCT \citep{chen2022effective} &  \xmark & \cmark & \xmark \\
STRIP \citep{gao2019strip} &  \cmark & \xmark & \xmark \\
AC \citep{chen2018detecting} &\cmark & \cmark & \xmark \\
SS \citep{tran2018spectral} &\cmark & \cmark & \xmark  \\
SPECTRE \citep{hayase2021spectre} &\cmark & \cmark & \xmark\\
SCAn \citep{tang2021demon} &  \cmark & \xmark & \cmark\\
SPC \citep{guo2023scaleup} &   \cmark & \cmark & \xmark \\
CD \citep{huang2023distilling} &  \cmark & \xmark & \cmark \\
Meta-SIFT \citep{zeng2022sift} & \cmark & \cmark & \xmark \\
ASSET \citep{pan2023asset} & \cmark & \xmark & \cmark \\
\midrule
 Ours & \cmark & \cmark & \cmark \\
\bottomrule[1pt]
\end{tabular}
}
\vspace*{-4mm}
\end{wraptable}%%%

% Free of Clean Data Free of Detection Threshold

% To address the problem of identification of backdoor data points, we focus on an observation from various works in the literature: backdoor samples have a variety of different properties than clean samples (often called \textit{backdoor signatures}). Such properties are generally used directly for identification \textit{or} as an intermediate step for further training of models.
% % robust training of models. \RW{not sure what robust training means here? Adversarial training also belongs to robust training.}

A class of defense methods like ABL \citep{li2021anti}, DBD \citep{huang2022backdoor}, SD-FCT \citep{chen2022effective} roughly detect backdoor / clean samples, aimed at unlearning these samples for ease of downstream processing. However, it is worth noting that these methods are typically \textit{not} designed for the accurate identification of backdoor samples, often leading to large errors in backdoor data selection.
In \textbf{Table\,\ref{tab: realtable}}, we denote the suitability of each existing method for `Backdoor Data Identification' with a {\xmark} or {\cmark}.
%We indicate this by \xmark \  for 'Backdoor Data Identification' in \textbf{Table\,\ref{tab: realtable}}-column 2.

Several other methods are designed solely for backdoor identification. A group of such methods \citep{chen2018detecting,tang2021demon,hayase2021spectre,tran2018spectral} use separable latent representations learnt by a classifier to distinguish between backdoor and clean data points. This strong \textit{latent separability assumption} was challenged through adaptive attacks in \citep{qi2023revisiting}, which mitigates this separation and subsequently, such identification methods. Moreover, we observe that they violate either \textbf{P1} or \textbf{P2} (\textbf{Table\,\ref{tab: realtable}}-row 5-8). Recent backdoor identification methods like \citep{guo2023scaleup,huang2023distilling} rely on different backdoor signatures like the scaled prediction consistency and cognitive pattern. However, the former requires setting a manual detection threshold while the latter requires clean samples for accurate identification. 

To the best of our knowledge, the most relevant work to ours is \citep{zeng2022sift}, which aims to sift out clean samples from a backdoored dataset. The work develops a bi-level optimization formulation that aims to minimise the cross-entropy loss of clean samples while simultaneously maximise that for backdoor samples. However, we note that this selection of clean data \textit{is a relatively easier problem} because of its abundance when compared to backdoor data. For example, let us consider a dataset with 45k clean samples and 5k backdoor samples. Precisely, the error rate of identifying 10\% clean samples out of 45k will be much lower when compared to that of identifying 10\% backdoor samples out of 5k; see more discussions in Section \ref{sec: expres}. Moreover, this method requires the prior knowledge of the poisoning ratio for accurately identifying all backdoor samples thus violating \textbf{P2}.

% Scores obtained from defense methods like ABL \citep{li2021anti}, DBD \citep{huang2022backdoor} are used for downstream processing such as unlearning the backdoor samples. The sample distinguishment (SD) module from \cite{chen2022effective} uses the FCT metric to distinguish between backdoor and clean samples, however these are used downstream for backdoor removal via unlearning or secure training via a modified contrastive learning. Thus, these methods are \textit{not} designed for accurate identification of backdoor samples , but can be easily used to do so. We indicate this as 'Applicability of Identification' in \textbf{Table\,\ref{tab: realtable}}-column 2.

% A class of backdoor identification methods use separable latent representations learnt by a classifier to distinguish between backdoor and clean data points. Specifically, identification is achieved by \textit{separation} in silhoutte scores of clusters from network activations \citep{chen2018detecting}, or robust statistics \citep{tang2021demon,hayase2021spectre} or outlier scores computed using centered representation matrix and its top singular vector \citep{tran2018spectral}. While \cite{tang2021demon} assumes the presence of clean data, the other works use heuristic methods to threshold the mentioned score vectors. Moreover, these methods fundamentally rely on the \textit{latent separability assumption} , which was challenged through adaptive attacks in \cite{qi2023revisiting}, which mitigates this separation and subsequently, such identification methods. 

Given this, existing backdoor identification methods do not satisfy either \textbf{P1} or \textbf{P2} of the studied practical condition setting. We have demonstrated this comprehensively in {Table\,\ref{tab: realtable}} and {Table\, \ref{tab: idtable}}. In our work, we leverage the scaled prediction consistency signature of backdoor data points to develop a novel backdoor identification method, which satisfies both of our practical conditions. 
\textbf{Our contributions} are summarized below:

% \ding{182} Reveal SPC failures

% \ding{183} Our method

% \ding{184} Experiment success
\ding{182} We peer into the usage of the scaled prediction consistency signature and provide various insights explaining its limitations. 
% Propelled by these insights, we develop a novel loss function called Mask-Aware SPC (MSPC).
% peer into the usage of the vanilla SPC (scaled prediction consistency) as a method for detecting backdoor attack fingerprints, and uncover its limitations in accurately identifying poisoned backdoor training samples. This motivates us to develop a new SPC method, referred to as Mask-Aware SPC (MSPC), which can be  effectively  applied to sifting out backdoor data samples. 

%We revisit the vanilla SPC method and find various limitations in this method as illustrated in Section \ref{sec: SPCRevisit}. 

\ding{183} Propelled by these insights, we develop a novel loss function called Mask-Aware SPC (MSPC). Using this loss, we develop a practical algorithm to identify backdoor samples, satisfying both $\textbf{P1}$ and $\textbf{P2}$. Our algorithm treats the problem as a hierarchical data-splitting task and optimizes the  MSPC loss function using bi-level optimization techniques.

% We introduce a practical algorithm called Scale-SIFT in Section \ref{sec: Scalesift}, which proposes to solve this problem by setting up as a hierarchical data-splitting problem, which aims to optimize an advanced SPC-based loss function via bi-level optimization. We do not set a threshold in an heuristic manner nor do we use any clean samples to do so.

\ding{184} Lastly, we conduct a comprehensive evaluation of our method across a range of metrics, including AUROC and TPR/FPR, against a wide variety of backdoor attacks. This includes basic BadNets, CleanLabel attacks, and more sophisticated Warping-based backdoor attacks at various poisoning rates and across different datasets.  Our approach often outperforms or performs at par with baseline methods (which {do not} satisfy the practical constraints).

\section{Related Work}\label{sec: relatedwork}
\textbf{Backdoor attacks.} Backdoor attacks aim to inject the backdoor trigger to the target model, so that the target model will consistently misclassify the backdoor samples with the backdoor trigger to the target label and behave normally on the clean samples. 
% Backdoor attacks can be applied under different threat models. 
Based on the knowledge of the adversaries, the backdoor attacks can be categorized into two tracks: data poisoning based attacks \citep{gu2017badnets,liu2018trojaning,chen2017targeted,turner2019label,zhao2020clean,nguyen2021wanet,li2021invisible,taneja2022does} and training manipulation based attacks \citep{garg2020can,lin2020composite,shumailov2021manipulating,bagdasaryan2021blind, tang2020embarrassingly, doan2021backdoor}. 
% The data poisoning based attacks can manipulate the datasets including data samples and/or labels without changing the training pipelines while the training manipulation based attacks can change the training process, manipulate the model parameters, or even add malicious modules directly. 
This paper will focus on the first category, the data poisoning based attacks. Attack methods such as BadNets \citep{gu2017badnets}, Trojan \citep{liu2018trojaning}, and Blend \citep{chen2017targeted} add simple trigger patterns like square patches or blend another figure into the background of backdoor samples, then mislabel them to the target class. 
% Such attacks, although effective, have different target labels from the ground truth. 
The clean-label attack \citep{turner2019label} is designed to poison only the target-class samples through adversarial attacks, so that the backdoor trigger injection avoids label manipulation and becomes more stealthy. Further, TUAP \citep{zhao2020clean} incorporates optimized universal adversarial perturbation into the clean-label attack and improves the attack success rate. Later, more invisible and sample-specific methods like WaNet \citep{nguyen2021wanet} and ISSBA \citep{li2021invisible} are proposed to bypass the backdoor defenses that assume the backdoor trigger is sample-agnostic. 
% Our work will look into the intrinsic common feature of these attacks' triggers. \SL{[I did not answer the last sentence.]}

\textbf{Backdoor defenses beyond identification.} 
% \RW{Section 1 talks about a lot of related works on backdoor sample detection. Maybe name this paragraph as Backdoor defenses beyond backdoor sample detection?}
To defend against backdoor attacks, numerous methods have been developed and can be divided into different categories. Backdoor trigger recovery \citep{wang2019neural,guo2019tabor,liu2019abs,sun2020poisoned,liu2022complex,xiang2022post,hu2021trigger} aims to synthesize the backdoor trigger used by the adversary while backdoor model reconstruction \citep{borgnia2021strong,huang2022backdoor,li2021anti,pal2023towards} attempts to purify the backdoor model by eliminating the backdoor effect. Backdoor model detection \citep{chen2019deepinspect,kolouri2020universal,wang2020practical,shen2021backdoor,xu2021detecting} identifies whether a model is poisoned from training on backdoor samples. However, the focal point of our work is identifying backdoor samples from a given dataset. We have elucidated various backdoor identification methods in {Table \ref{tab: realtable}}. As mentioned previously, these methods do not meet our \textit{practical constraints}, which we will address in this paper.
\section{Preliminaries and Problem Setup}\label{sec: setup}

\paragraph{Backdoor attacks and defender’s capabilities.}  We consider the setting where the user receives a backdoor dataset from a third party or external source and can train their model using that dataset. Given a clean dataset $\mathcal{D} = \{( \mathbf{x}_i, y_i)\}_{i=1}^N = \mathcal{D}_m \bigcup \mathcal{D}_n$ and a poisoning ratio $\gamma = \frac{|\mathcal{D}_m|}{|\mathcal{D}|}$ (with $\gamma \ll 1$), the adversary injects the   backdoor data $\mathcal{D}_b$ in place of $\mathcal{D}_m$. $\mathcal{D}_b$ is generated using a certain backdoor data generator $\mathcal{G}$ such that $\mathcal{D}_b = \{(\mathbf{x}', y_t) \ | \mathbf{x}'= \mathcal{G}(\mathbf{x}), (\mathbf{x},y) \in \mathcal{D}_m\}$, where $y_t \neq y$ is a target label. Thus, the user receives the backdoor dataset $\mathcal{D}_p = \mathcal{D}_b \bigcup \mathcal{D}_n$, and training on this leads to a backdoor model vulnerable to test-time manipulation by the adversary.

Let $\mathcal{F}_{\boldsymbol \theta} : \mathcal{X} \rightarrow \mathbb{R}^{C}$ denote the ML model that we consider, parameterized by $\boldsymbol \theta$. Here $\mathcal{X}$ is the input space and the model gives as an output a real vector with $C$ dimensions (\textit{i.e.}, $C$ classes). Hence, the predicted class of an input $\mathbf{x}$ by the model $\mathcal{F}_{\boldsymbol \theta}$ is given as $ \argmax_{[C]} \mathcal{F}_{\boldsymbol \theta}(\mathbf{x})$, where $[C] = \{1,2,3,...,C\}$. In the rest of the paper, we use the shorthand of this notation as $ \argmax\mathcal{F}_{\boldsymbol \theta}(\mathbf{x})$. 

% We consider the setting where the user gets data from an adversary which is potentially backdoored. \RW{this is a repeat sentence.} 
In the above setting, the defender wants to identify backdoor samples from this dataset under the \textbf{practical conditions} as mentioned in Section \ref{sec: intro}. In this premise, the defender is \textit{free to choose to train any model} on this dataset for backdoor identification.

\paragraph{Warm-up: SPC alone is not sufficient.}

Of the various backdoor signatures in the literature (see Section \ref{sec: intro}), we draw inspiration from the \textit{scale invariance signature} of backdoor samples, 
\textit{i.e.}, the SPC (scaled prediction consistency)  loss \citep{guo2023scaleup}. This is because: \textbf{(1)} Of the various signatures applicable for backdoor sifting, SPC satisfies the constraint of no prior clean data (\textbf{P1}); \textbf{(2)} SPC is computationally efficient; And \textbf{(3)} SPC does not rely on the {latent separability assumption}. 
Specifically, given an input datapoint $\mathbf{x}$ and a set of scales $\mathcal{S} = \{2,3,...,12\}$, the vanilla SPC loss measures agreement in the prediction of $\mathbf{x}$ and  that of scalar multiplication of $\mathbf{x}$: 

\vspace*{-5mm}
{\small{
\begin{align}
\ell_{\mathrm{SPC}} (\mathbf{x}) = \sum_{n \in \mathcal{S}}\frac{\mathbbm{1}(\displaystyle \argmax\mathcal{F}_{\boldsymbol \theta}(\mathbf{x}) = \displaystyle \argmax\mathcal{F}_{\boldsymbol \theta}(n \cdot\mathbf{x}))}{|\mathcal{S}|},
\label{eq: SPC}
\end{align}
}}%
\begin{wrapfigure}{r}{52mm}
%\vspace*{-9mm}
\vspace*{-5mm}
\centerline{
\includegraphics[width=52mm,height=!]{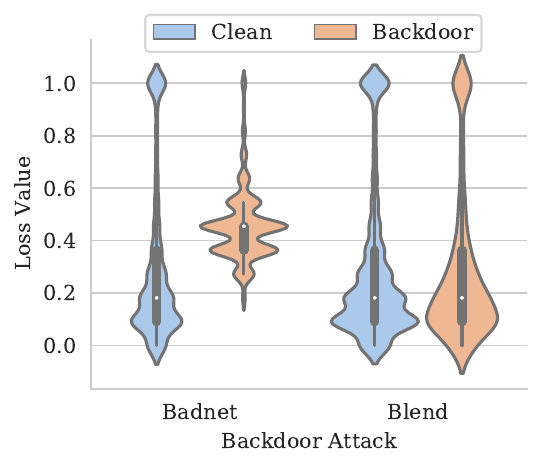}
}
\vspace*{-3mm}
\caption{\footnotesize{Violin plots for SPC loss for backdoor poisoned data and clean data when facing Badnet attack   \citep{gu2017badnets} and Blended  attack \citep{chen2017targeted}. 
}}
  \label{fig: SPCfailure}
 \vspace*{-5mm}
\end{wrapfigure}%%%%
where $\mathbbm{1}$ denotes the indicator function, $|\mathcal{S}|$ denotes the cardinality of the set of scales $\mathcal{S}$, 
% \RW{$|\cdot|$ has been used before} 
and $n$ is a scaling constant. The work \citep{guo2023scaleup} showed that for backdoor samples, the predicted class of a particular sample remains the same even when multiplied by scales,   due to the strong correlation established between the trigger and the target label. Hence, for such a backdoor sample and for a particular scale $n \in \mathcal{S}$, the indicator function in  \eqref{eq: SPC} gives a value close to $+1$, which is averaged over the number of scales.

Though the vanilla SPC loss has been shown to perform well in detecting backdoor samples over various attacks, we demonstrate some issues via \textbf{Figure\,\ref{fig: SPCfailure}}. \textbf{(1)} For Badnet attack \citep{gu2017badnets}, while the clean (or backdoor) samples have low (or high) SPC loss on average, we observe that there is a large variance in the loss values for both clean and backdoor samples. \textbf{(2)} The loss value distribution is indistinguishable for the Blended attack \citep{chen2017targeted}. \textbf{(3)} Moreover, to distinguish between clean and backdoor samples, the defender needs to set a threshold on  SPC values (thus violating \textbf{P2}). 
Given the pros and cons of this signature, we ask:
\begin{center}
% \vspace{-3mm}
    \textit{(\textbf{Q}) Can we exploit the advantages of the scale invariance signature to develop a backdoor identification algorithm, which simultaneously satisfy \textbf{P1} and \textbf{P2} as introduced in Sec.\,\ref{sec: intro}?}
\end{center}

\section{Proposed Method}
\label{sec: PropMethod}

\begin{figure*}[t]
  \centering
  \begin{adjustbox}{max width=1\textwidth,bgcolor=verylightgray}
  \begin{tabular}{@{\hskip 0.00in}c@{\hskip 0.00in} @{\hskip -0.07in}c @{\hskip -0.07in} | @{\hskip -0.07in}  c @{\hskip -0.07in}
  }
& 
\colorbox{lightgray}{ \textbf{CIFAR-10}}
&
\colorbox{lightgray}{ \textbf{Imagenet}} 
\\
 \begin{tabular}{c} %{@{}c@{}}  
\rotatebox{90}{\parbox{12.9em}{\centering  \colorbox{lightgray}{\textbf{Insight 1}}}}
  \\
 \hline
 \\
\rotatebox{90}{\parbox{18em}{\centering  \colorbox{lightgray}{\textbf{Insight 2}}}}
\end{tabular} 
& 
 % CIFAR10 INSIGHT 1
\begin{tabular}{ccccc}
 % CIFAR 10 Insight 2
 \begin{tabular}{cccccc}
  \begin{tabular}{@{\hskip 0.0in}c@{\hskip 0.0in} }
 \rotatebox{90}{\parbox{5em}{\centering \footnotesize{}}}
 \end{tabular}
 \begin{tabular}{@{\hskip 0.0in}c@{\hskip 0.0in} }
 \parbox{5em}{\centering \footnotesize{Original}}
 \end{tabular}
 \begin{tabular}{@{\hskip 0.0in}c@{\hskip 0.0in} }
 \parbox{5em}{\centering \footnotesize{$\times 5$}}
 \end{tabular}
 \begin{tabular}{@{\hskip 0.0in}c@{\hskip 0.0in} }
 \parbox{5em}{\centering \footnotesize{$\times 7$}}
 \end{tabular}
 \begin{tabular}{@{\hskip 0.0in}c@{\hskip 0.0in} }
 \parbox{5em}{\centering \footnotesize{$\times 9$}}
 \end{tabular}
 \begin{tabular}{@{\hskip 0.0in}c@{\hskip 0.0in} }
 \parbox{5em}{\centering \footnotesize{$\times 11$}}
 \end{tabular}
\\
 \begin{tabular}{@{\hskip 0.0in}c@{\hskip 0.0in} }
 \rotatebox{90}{\parbox{5em}{\centering \footnotesize{Blend}}}
 \end{tabular}
  \begin{tabular}{@{\hskip 0.0in}c@{\hskip 0.0in} }
 \parbox[c]{5em}{\includegraphics[width=5em]{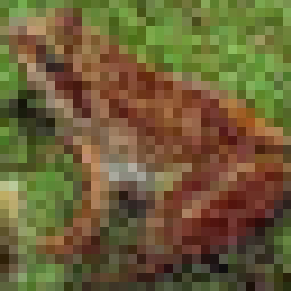}}\\ \colorbox{lightgreen}{Pred : 1}
  \end{tabular}
  \begin{tabular}{@{\hskip 0.0in}c@{\hskip 0.0in} }
 \parbox[c]{5em}{\includegraphics[width=5em]{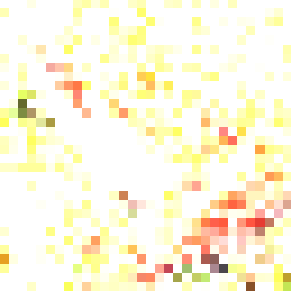}}\\ \colorbox{lightred}{Pred : 1}
  \end{tabular}
  \begin{tabular}{@{\hskip 0.0in}c@{\hskip 0.0in} }
 \parbox[c]{5em}{\includegraphics[width=5em]{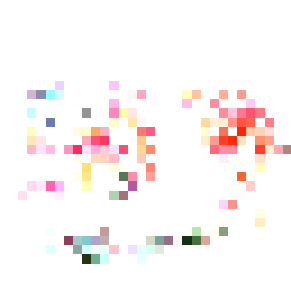}}\\\colorbox{lightred}{Pred : 1}
  \end{tabular}
  \begin{tabular}{@{\hskip 0.0in}c@{\hskip 0.0in} }
 \parbox[c]{5em}{\includegraphics[width=5em]{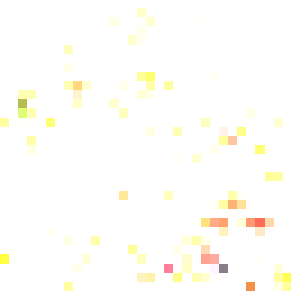}}\\\colorbox{lightred}{Pred : 1}
  \end{tabular}
  \begin{tabular}{@{\hskip 0.0in}c@{\hskip 0.0in} }
  \parbox[c]{5em}{\includegraphics[width=5em]{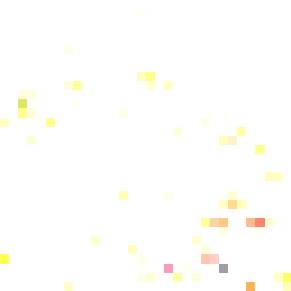}}\\\colorbox{lightred}{Pred : 1}
  \end{tabular}
  \\
  \begin{tabular}{@{\hskip 0.0in}c@{\hskip 0.0in} }
 \rotatebox{90}{\parbox{5em}{\centering \footnotesize{Trigger}}}
 \end{tabular}
  \begin{tabular}{@{\hskip 0.0in}c@{\hskip 0.0in} }
 \parbox[c]{5em}{\includegraphics[width=5em]{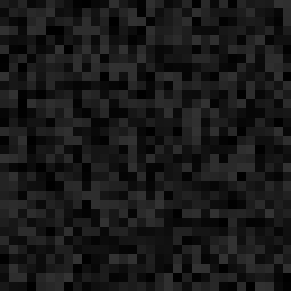}}\\{ }
  \end{tabular}
  \begin{tabular}{@{\hskip 0.0in}c@{\hskip 0.0in} }
 \parbox[c]{5em}{\includegraphics[width=5em]{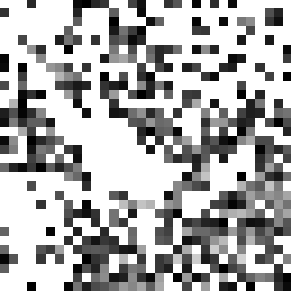}}\\{ }
  \end{tabular}
  \begin{tabular}{@{\hskip 0.0in}c@{\hskip 0.0in} }
 \parbox[c]{5em}{\includegraphics[width=5em]{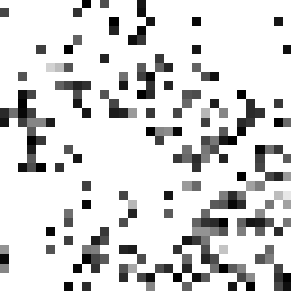}}\\{ }
  \end{tabular}
  \begin{tabular}{@{\hskip 0.0in}c@{\hskip 0.0in} }
 \parbox[c]{5em}{\includegraphics[width=5em]{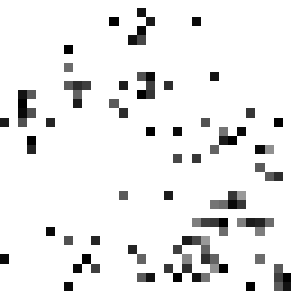}}\\{ }
  \end{tabular}
  \begin{tabular}{@{\hskip 0.0in}c@{\hskip 0.0in} }
  \parbox[c]{5em}{\includegraphics[width=5em]{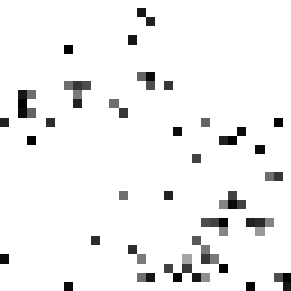}}\\{ }
  \end{tabular}
  \vspace{1mm}
 \\
 \hline
 \\
 \begin{tabular}{@{\hskip 0.0in}c@{\hskip 0.0in} }
 \rotatebox{90}{\parbox{5em}{\centering \footnotesize{Scenario 1}}}
 \end{tabular}
  \begin{tabular}{@{\hskip 0.0in}c@{\hskip 0.0in} }
 \parbox[c]{5em}{\includegraphics[width=5em]{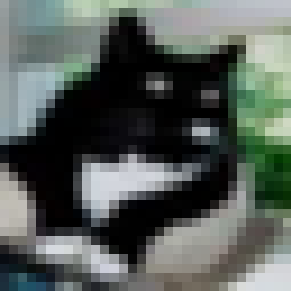}}
 \\\colorbox{lightgreen}{Pred : 3}
  \end{tabular}
  \begin{tabular}{@{\hskip 0.0in}c@{\hskip 0.0in} }
 \parbox[c]{5em}{\includegraphics[width=5em]{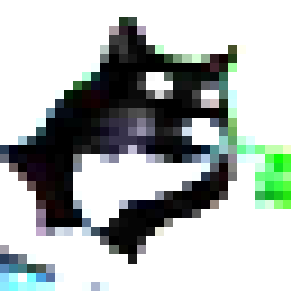}}\\\colorbox{lightred}{Pred : 3}
  \end{tabular}
  \begin{tabular}{@{\hskip 0.0in}c@{\hskip 0.0in} }
 \parbox[c]{5em}{\includegraphics[width=5em]{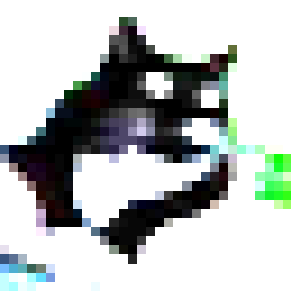}}\\\colorbox{lightred}{Pred : 3}
  \end{tabular}
  \begin{tabular}{@{\hskip 0.0in}c@{\hskip 0.0in} }
 \parbox[c]{5em}{\includegraphics[width=5em]{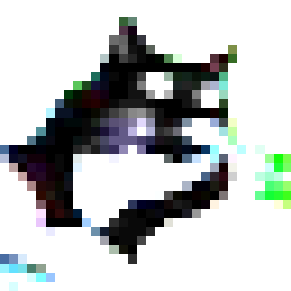}}\\\colorbox{lightred}{Pred : 3}
  \end{tabular}
  \begin{tabular}{@{\hskip 0.0in}c@{\hskip 0.0in} }
  \parbox[c]{5em}{\includegraphics[width=5em]{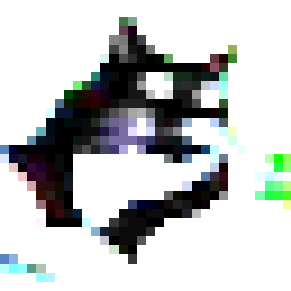}}\\\colorbox{lightred}{Pred : 3}
  \end{tabular}
  \\
  \begin{tabular}{@{\hskip 0.0in}c@{\hskip 0.0in} }
 \rotatebox{90}{\parbox{5em}{\centering \footnotesize{Scenario 2}}}
 \end{tabular}
  \begin{tabular}{@{\hskip 0.0in}c@{\hskip 0.0in} }
 \parbox[c]{5em}{\includegraphics[width=5em]{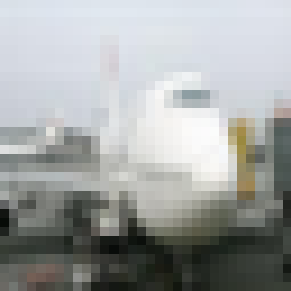}}\\\colorbox{lightgreen}{Pred : 0}
  \end{tabular}
  \begin{tabular}{@{\hskip 0.0in}c@{\hskip 0.0in} }
 \parbox[c]{5em}{\includegraphics[width=5em]{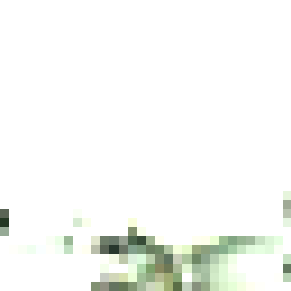}}\\\colorbox{lightred}{Pred : 0}
  \end{tabular}
  \begin{tabular}{@{\hskip 0.0in}c@{\hskip 0.0in} }
 \parbox[c]{5em}{\includegraphics[width=5em]{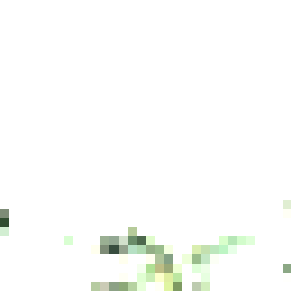}}\\\colorbox{lightred}{Pred : 0}
  \end{tabular}
  \begin{tabular}{@{\hskip 0.0in}c@{\hskip 0.0in} }
 \parbox[c]{5em}{\includegraphics[width=5em]{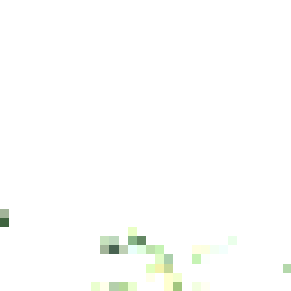}}\\\colorbox{lightred}{Pred : 0}
  \end{tabular}
  \begin{tabular}{@{\hskip 0.0in}c@{\hskip 0.0in} }
  \parbox[c]{5em}{\includegraphics[width=5em]{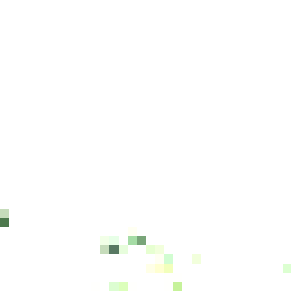}}\\\colorbox{lightred}{Pred : 0}
  \end{tabular}
  \\
  \begin{tabular}{@{\hskip 0.0in}c@{\hskip 0.0in} }
 \rotatebox{90}{\parbox{5em}{\centering \footnotesize{Scenario 3}}}
 \end{tabular}
  \begin{tabular}{@{\hskip 0.0in}c@{\hskip 0.0in} }
 \parbox[c]{5em}{\includegraphics[width=5em]{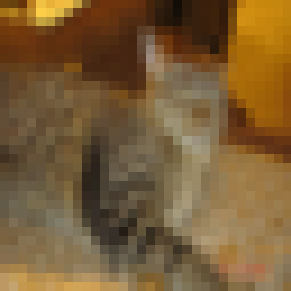}}\\\colorbox{lightgreen}{Pred : 3}
  \end{tabular}
  \begin{tabular}{@{\hskip 0.0in}c@{\hskip 0.0in} }
 \parbox[c]{5em}{\includegraphics[width=5em]{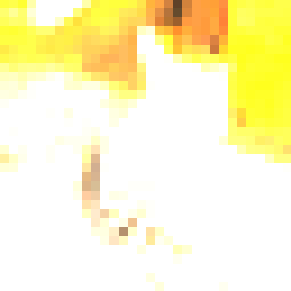}}\\\colorbox{lightred}{Pred : 3}
  \end{tabular}
  \begin{tabular}{@{\hskip 0.0in}c@{\hskip 0.0in} }
 \parbox[c]{5em}{\includegraphics[width=5em]{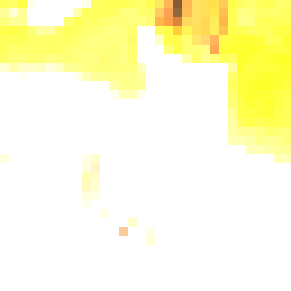}}\\\colorbox{lightred}{Pred : 3}
  \end{tabular}
  \begin{tabular}{@{\hskip 0.0in}c@{\hskip 0.0in} }
 \parbox[c]{5em}{\includegraphics[width=5em]{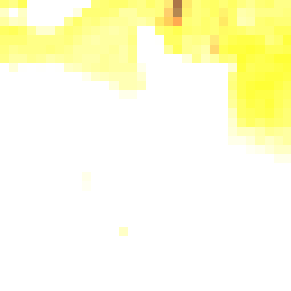}}\\\colorbox{lightred}{Pred : 3}
  \end{tabular}
  \begin{tabular}{@{\hskip 0.0in}c@{\hskip 0.0in} }
  \parbox[c]{5em}{\includegraphics[width=5em]{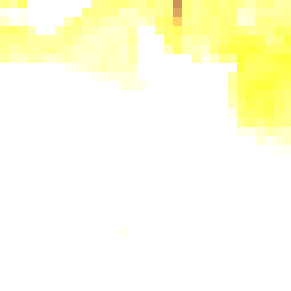}}\\\colorbox{lightgreen}{Pred : 0}
  \end{tabular}
\\
 \end{tabular}
\end{tabular}
& 
% IMAGENET INSIGHT 1
\begin{tabular}{c}
   \begin{tabular}{ccccc}
   \begin{tabular}{@{\hskip 0.0in}c@{\hskip 0.0in} }
 \rotatebox{90}{\parbox{5em}{\centering \footnotesize{}}}
 \end{tabular}
    \begin{tabular}{@{\hskip 0.0in}c@{\hskip 0.0in} }
 \parbox{5em}{\centering \footnotesize{Original}}
 \end{tabular}
 \begin{tabular}{@{\hskip 0.0in}c@{\hskip 0.0in} }
 \parbox{5em}{\centering \footnotesize{$\times 5$}}
 \end{tabular}
 \begin{tabular}{@{\hskip 0.0in}c@{\hskip 0.0in} }
 \parbox{5em}{\centering \footnotesize{$\times 7$}}
 \end{tabular}
 \begin{tabular}{@{\hskip 0.0in}c@{\hskip 0.0in} }
 \parbox{5em}{\centering \footnotesize{$\times 9$}}
 \end{tabular}
 \begin{tabular}{@{\hskip 0.0in}c@{\hskip 0.0in} }
 \parbox{5em}{\centering \footnotesize{$\times 11$}}
 \end{tabular}
 \\
 \begin{tabular}{@{\hskip 0.0in}c@{\hskip 0.0in} }
 \rotatebox{90}{\parbox{5em}{\centering \footnotesize{Badnet}}}
 \end{tabular}
  \begin{tabular}{@{\hskip 0.0in}c@{\hskip 0.0in} }
 \parbox[c]{5em}{\includegraphics[width=5em]{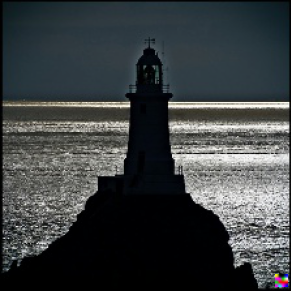}}\\ \colorbox{lightgreen}{Pred : 1}
  \end{tabular}
  \begin{tabular}{@{\hskip 0.0in}c@{\hskip 0.0in} }
 \parbox[c]{5em}{\includegraphics[width=5em]{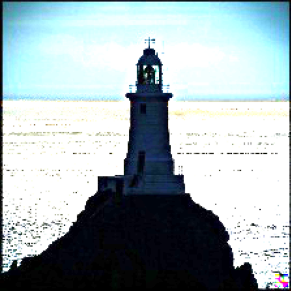}}\\ \colorbox{lightred}{Pred : 71}
  \end{tabular}
  \begin{tabular}{@{\hskip 0.0in}c@{\hskip 0.0in} }
 \parbox[c]{5em}{\includegraphics[width=5em]{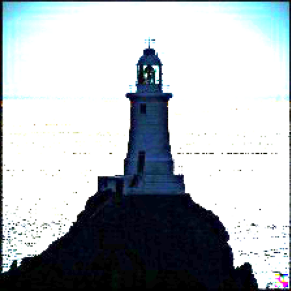}}\\\colorbox{lightred}{Pred : 71}
  \end{tabular}
  \begin{tabular}{@{\hskip 0.0in}c@{\hskip 0.0in} }
 \parbox[c]{5em}{\includegraphics[width=5em]{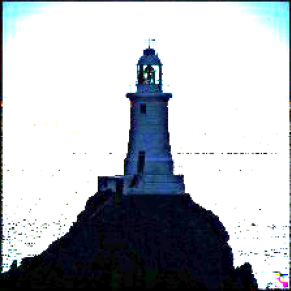}}\\\colorbox{lightred}{Pred : 71}
  \end{tabular}
  \begin{tabular}{@{\hskip 0.0in}c@{\hskip 0.0in} }
  \parbox[c]{5em}{\includegraphics[width=5em]{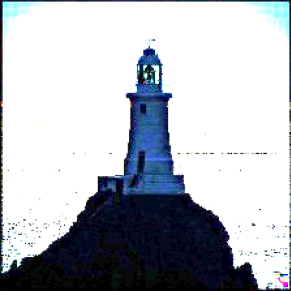}}\\\colorbox{lightred}{Pred : 71}
  \end{tabular}
  \\
  \begin{tabular}{@{\hskip 0.0in}c@{\hskip 0.0in} }
 \rotatebox{90}{\parbox{5em}{\centering \footnotesize{Blend}}}
 \end{tabular}
  \begin{tabular}{@{\hskip 0.0in}c@{\hskip 0.0in} }
 \parbox[c]{5em}{\includegraphics[width=5em]{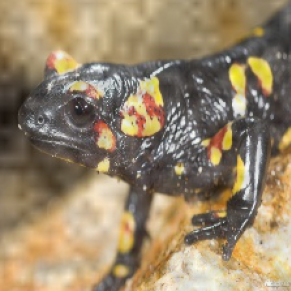}}\\\colorbox{lightgreen}{Pred : 1}
  \end{tabular}
  \begin{tabular}{@{\hskip 0.0in}c@{\hskip 0.0in} }
 \parbox[c]{5em}{\includegraphics[width=5em]{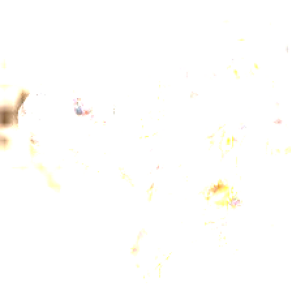}}\\\colorbox{lightred}{Pred : 159}
  \end{tabular}
  \begin{tabular}{@{\hskip 0.0in}c@{\hskip 0.0in} }
 \parbox[c]{5em}{\includegraphics[width=5em]{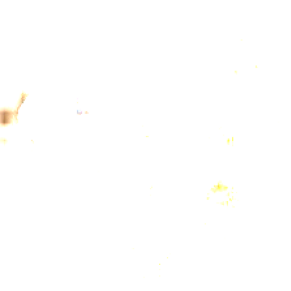}}\\\colorbox{lightred}{Pred : 159}
  \end{tabular}
  \begin{tabular}{@{\hskip 0.0in}c@{\hskip 0.0in} }
 \parbox[c]{5em}{\includegraphics[width=5em]{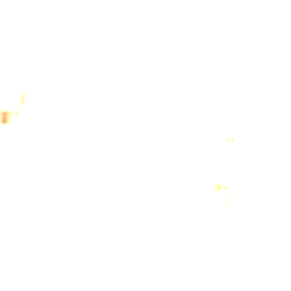}}\\\colorbox{lightred}{Pred : 159}
  \end{tabular}
  \begin{tabular}{@{\hskip 0.0in}c@{\hskip 0.0in} }
  \parbox[c]{5em}{\includegraphics[width=5em]{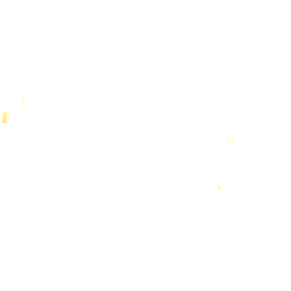}}\\\colorbox{lightred}{Pred : 159}
  \end{tabular}
 \\
 \hline
 % IMAGENET INSIGHT 2
 \\
 \begin{tabular}{@{\hskip 0.0in}c@{\hskip 0.0in} }
 \rotatebox{90}{\parbox{5em}{\centering \footnotesize{Scenario 1}}}
 \end{tabular}
  \begin{tabular}{@{\hskip 0.0in}c@{\hskip 0.0in} }
 \parbox[c]{5em}{\includegraphics[width=5em]{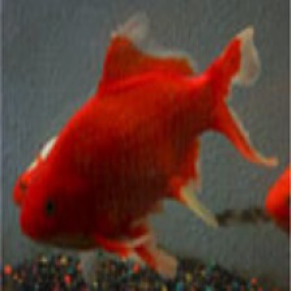}}\\\colorbox{lightgreen}{Pred : 0}
  \end{tabular}
  \begin{tabular}{@{\hskip 0.0in}c@{\hskip 0.0in} }
 \parbox[c]{5em}{\includegraphics[width=5em]{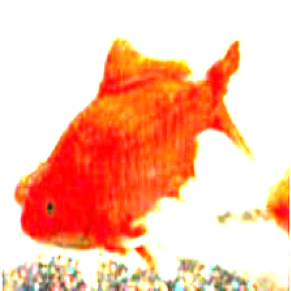}}\\\colorbox{lightred}{Pred : 0}
  \end{tabular}
  \begin{tabular}{@{\hskip 0.0in}c@{\hskip 0.0in} }
 \parbox[c]{5em}{\includegraphics[width=5em]{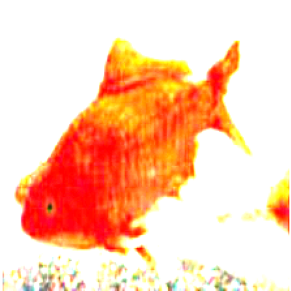}}\\\colorbox{lightred}{Pred : 0}
  \end{tabular}
  \begin{tabular}{@{\hskip 0.0in}c@{\hskip 0.0in} }
 \parbox[c]{5em}{\includegraphics[width=5em]{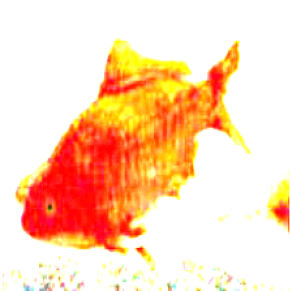}}\\\colorbox{lightred}{Pred : 0}
  \end{tabular}
  \begin{tabular}{@{\hskip 0.0in}c@{\hskip 0.0in} }
  \parbox[c]{5em}{\includegraphics[width=5em]{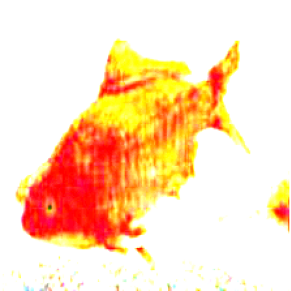}}\\\colorbox{lightred}{Pred : 0}
  \end{tabular}
  \end{tabular}
 \\
 \begin{tabular}{@{\hskip 0.0in}c@{\hskip 0.0in} }
 \rotatebox{90}{\parbox{5em}{\centering \footnotesize{Scenario 1}}}
 \end{tabular}
  \begin{tabular}{@{\hskip 0.0in}c@{\hskip 0.0in} }
 \parbox[c]{5em}{\includegraphics[width=5em]{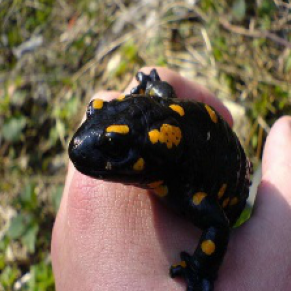}}\\\colorbox{lightgreen}{Pred : 1}
  \end{tabular}
  \begin{tabular}{@{\hskip 0.0in}c@{\hskip 0.0in} }
 \parbox[c]{5em}{\includegraphics[width=5em]{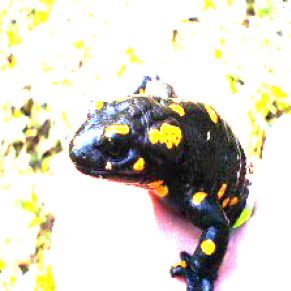}}\\\colorbox{lightred}{Pred : 1}
  \end{tabular}
  \begin{tabular}{@{\hskip 0.0in}c@{\hskip 0.0in} }
 \parbox[c]{5em}{\includegraphics[width=5em]{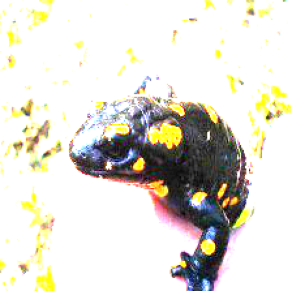}}\\\colorbox{lightred}{Pred : 1}
  \end{tabular}
  \begin{tabular}{@{\hskip 0.0in}c@{\hskip 0.0in} }
 \parbox[c]{5em}{\includegraphics[width=5em]{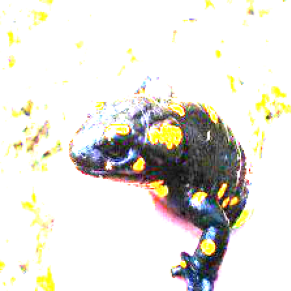}}\\\colorbox{lightred}{Pred : 1}
  \end{tabular}
  \begin{tabular}{@{\hskip 0.0in}c@{\hskip 0.0in} }
  \parbox[c]{5em}{\includegraphics[width=5em]{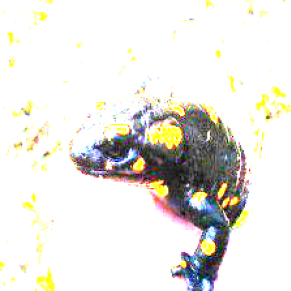}}\\\colorbox{lightred}{Pred : 1}
  \end{tabular}
 \\
 \begin{tabular}{@{\hskip 0.0in}c@{\hskip 0.0in} }
 \rotatebox{90}{\parbox{5em}{\centering \footnotesize{Scenario 1}}}
 \end{tabular}
  \begin{tabular}{@{\hskip 0.0in}c@{\hskip 0.0in} }
 \parbox[c]{5em}{\includegraphics[width=5em]{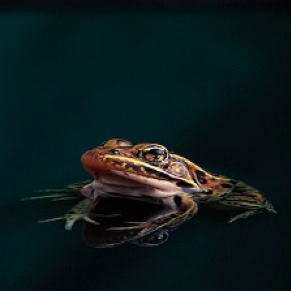}}\\\colorbox{lightgreen}{Pred : 3}
  \end{tabular}
  \begin{tabular}{@{\hskip 0.0in}c@{\hskip 0.0in} }
 \parbox[c]{5em}{\includegraphics[width=5em]{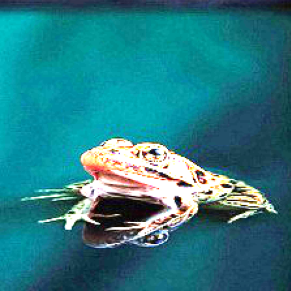}}\\\colorbox{lightred}{Pred : 3}
  \end{tabular}
  \begin{tabular}{@{\hskip 0.0in}c@{\hskip 0.0in} }
 \parbox[c]{5em}{\includegraphics[width=5em]{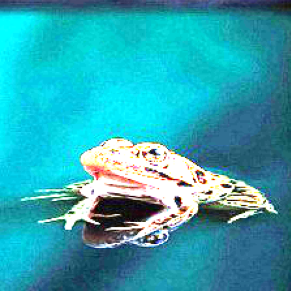}}\\\colorbox{lightred}{Pred : 3}
  \end{tabular}
  \begin{tabular}{@{\hskip 0.0in}c@{\hskip 0.0in} }
 \parbox[c]{5em}{\includegraphics[width=5em]{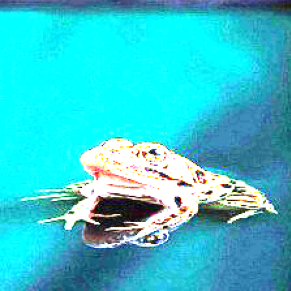}}\\\colorbox{lightred}{Pred : 3}
  \end{tabular}
  \begin{tabular}{@{\hskip 0.0in}c@{\hskip 0.0in} }
  \parbox[c]{5em}{\includegraphics[width=5em]{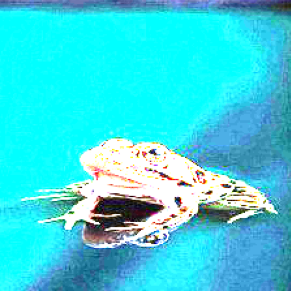}}\\\colorbox{lightred}{Pred : 3}
  \end{tabular}
\end{tabular}
\end{tabular}
  \end{adjustbox}
\caption{\footnotesize{Illustration of the various SPC limitations in terms of the elucidated insights. $\times a$ indicates an image is multipled by scalar $a$ and then constrained between 0 and 1. \textbf{Insight 1} : Backdoor samples with low SPC loss. \textbf{Insight 2} : Clean samples with low SPC loss. Predictions are marked as \setlength{\fboxsep}{0pt}\colorbox{lightgreen}{green} if it is an expected prediction, while it is marked \setlength{\fboxsep}{0pt}\colorbox{lightred}{red} for an unexpected prediction. Note that for backdoored samples, the expected label is 1 which is the target label.
}}
\label{fig: spclimit}
 \vspace*{-5mm}
\end{figure*}

\paragraph{Exploring SPC limitations: Insights gained. }
First, we delve into the vanilla SPC loss to gain insights from its various failure cases. 
%
% From \eqref{eq: SPC}, we expect the backdoor samples to have a high SPC loss because of the scale-invariant property of backdoors. Here we explore cases where backdoor samples obtain low SPC loss. We observe examples of this case in \textbf{Figure \ref{fig: spclimit}} - row 1 and 2 for both CIFAR-10 and ImageNet. \ding{182} Because of constraining the scaled pixel values to be between $[0,1]$, the trigger may mostly vanish when multiplied with a large scalar $n \in \mathcal{S}$. See the first two rows under 'CIFAR-10' : `Backdoor image $\mathbf x$' and Trigger $\mathbf t$ of \textbf{Figure\,\ref{fig: spclimit}} for examples. \ding{183} Additionally, the trigger may be 'blend' in with the background as seen in row 2 (ImageNet) of \textbf{Figure\,\ref{fig: spclimit}}. In higher scales, the trigger blends in the background of the sea.
%
From \eqref{eq: SPC}, backdoor samples are anticipated to exhibit high SPC loss values due to the scale-invariant nature of backdoors. However, instances where backdoor samples manifest low SPC loss values are explored, as illustrated in \textbf{Figure \ref{fig: spclimit}}. Two reasons for this anomaly are identified: 
% Firstly, the restriction of scaled pixel values to the range $[0,1]$, causing the trigger to vanish when multiplied with a large scalar 
Firstly, clipping pixel values to the range $[0,1]$ after  multiplying with a large scalar value causes the trigger to vanish (\textbf{Figure \ref{fig: spclimit}}-CIFAR10 row 1, 2 and ImageNet row 2); 
% \SL{[The following sentence is ill! I noticed several unprofessional writings, e.g., ImageNet. ]}
Secondly, the trigger may blend with the background at higher scales (\textbf{Figure \ref{fig: spclimit}}- ImageNet row 1). 
% \SL{[The previous sentence is ill. I did not follow.]}
%, especially observable in ImageNet examples. 
We have demonstrated the vanishing of the effective parts of the Blend trigger with CIFAR-10. As seen in \textbf{Figure \ref{fig: spclimit}}-CIFAR10 row 2, the trigger practically vanishes at $\times 7$, thus rendering the backdoor ineffective.

\begin{center}
\vspace*{-2mm}
	\setlength\fboxrule{0.1pt}
	\noindent\fcolorbox{black}[rgb]{0.97,0.97,0.97}{\begin{minipage}{0.96\columnwidth}
			%\begin{center}
				\vspace{-0.00cm}
	\textbf{Insight 1.} \textit{Backdoor samples} can obtain \textit{low} SPC loss because the backdoor pixels vanish or blend with the background when multiplied with higher scales. 
				\vspace{-0.00cm}
		%	\end{center}
	\end{minipage}}
	\vspace*{-1.5mm}
\end{center}

% We now consider clean samples with high SPC losses (which is contrary to expected behaviour). We observe that one of the most prevalent cases for this is where predictive features or the object of interest remain intact in spite of scaling. This is mainly possible because of having low pixel values constituting the object. For example, in row 3, column 'CIFAR-10' of \textbf{Figure\,\ref{fig: spclimit}}, the object cat has very low pixel values and stays intact even when multiplied with a scale of 11. See row 3, 4, column 'ImageNet' for more such examples. We also observe that in row 5 column 'ImageNet' of \textbf{Figure\,\ref{fig: spclimit}}, the object 'frog' stays intact because of its constrast with the background. We call this \textbf{Scenario 1} when considering clean samples with high SPC losses.

In contrast, we also examine instances of clean samples with high SPC losses, which deviate from expectations. This occurrence is prevalent when objects of interest maintain their structure despite scaling, primarily due to their low pixel values, as exemplified in \textbf{Figure\,\ref{fig: spclimit}}. 
% We call this \textbf{Scenario 1} when considering clean samples with high SPC losses.
% \SL{[Do we really need calling Scenario 1, 2, 3? It is strange as Scenario 1 corresponds to Insight 2??]}

\begin{center}
\vspace*{-2mm}
	\setlength\fboxrule{0.1pt}
	\noindent\fcolorbox{black}[rgb]{0.97,0.97,0.97}{\begin{minipage}{0.96\columnwidth}
			%\begin{center}
				\vspace{-0.00cm}
	\textbf{Insight 2.} \textit{Clean samples} can obtain \textit{high} SPC loss value because predictive features of the main object remain intact even at high scales.
				\vspace{-0.00cm}
		%	\end{center}
	\end{minipage}}
	\vspace*{-1.5mm}
\end{center}
% We observe two other scenarios, though less prevalent when considering clean samples with high SPC loss. As shown in \textbf{Figure\,\ref{fig: spclimit}}-row 2 column CIFAR-10, because of scaling, some images often completely vanish. That is an expected behavior. However, the caveat is that it produces a high SPC loss when the original class of the image is the same as the class predicted by the neural network when it has a completely white image (\textbf{Scenario 2}). Finally, we discover an intriguing phenomenon similar to the effect of spurious correlation \citep{sagawa2020investigation}: Certain spurious image patterns can be found simply when scaling the images by scalars. The predictions of images remain intact because of their reliance on such spurious features, as shown in \textbf{Figure\,\ref{fig: spclimit}}-row 5 column CIFAR-10  (\textbf{Scenario 3}). We note that this is not a systematic detection of spurious features, rather an observation which explains high SPC values for benign samples. We refer readers to Appendix \ref{app: ??} for a more detailed discussion.

Additionally, two less common scenarios are noted: Images completely vanish due to scaling, resulting in high SPC losses when the original class remains consistent 
% (\textbf{Scenario 2})
(\textbf{Figure \ref{fig: spclimit}}-CIFAR10 row 4), and the emergence of certain spurious correlations when scaling, which maintain image predictions 
% (\textbf{Scenario 3}). 
(\textbf{Figure \ref{fig: spclimit}}-CIFAR10 row 5).
% \SL{[Do we really need calling Scenario 2, 3?]}
This observation, detailed further in Appendix \ref{app: scen3}, elucidates high SPC values for benign samples without systematically detecting spurious features.

\paragraph{Enhancing SPC: Mask-aware SPC (MSPC).} 
Informed by the previous insights, we introduce a novel mask-aware SPC loss. Given a (known) mask $\mathbf{m}$  and a subtle linear shift $\tau$, we define the proposed MSPC loss below:

\vspace*{-5mm}
{\small{
\begin{align}
\begin{split}
    \ell_\mathrm{MSPC} (\mathbf{x}_i, \mathbf{m},  \tau) &= \frac{1}{|S|} \displaystyle \sum_{n \in S} \phi ( \argmax\mathcal{F}_{\boldsymbol \theta}(\mathbf{x}_i)  -\argmax\mathcal{F}_{\boldsymbol \theta}(n \cdot\mathbf{x}_i^m) ) , \\
    \mathbf{x}_i^m &= (\mathbf{x}_i - \tau) \odot \mathbf{m} , 
    %\\%+  (1 - \mathbf{m}) \odot \delta \\
  %  \ 
  ~~  \phi (x)  = \begin{cases}
                                1 & \text{if } x=0\\
                                -1 & \text{if } x \neq 0.
                            \end{cases}\\
\end{split}
\label{eq: MSPC}
\end{align}
}}%
In \eqref{eq: MSPC}, we consider a known mask $\mathbf m$ applied on an input $\mathbf{x}_i$, which is shifted by $\tau$ to form the masked image $\mathbf{x}^m_i = (\mathbf x_i - \tau) \odot \mathbf m$. The linear shift $\tau$ is to slightly shift the histogram of the trigger to lower pixel values. This can help in preserving the trigger at higher scales. We \textit{assume} that the mask $\mathbf m$ has encoded the focused `effective part' of the trigger in backdoor samples. This is reminiscent of \citep{huang2023distilling}, which proposed to extract the `minimal essence' of an image for prediction, \textit{i.e.}, the minimal mask such that the prediction of the model stays the same. 

\begin{figure*}[t]
  \centering
  \begin{adjustbox}{max width=1\textwidth,bgcolor=verylightgray}
  \begin{tabular}{@{\hskip 0.00in}c@{\hskip 0.00in} @{\hskip -0.07in}c @{\hskip -0.07in} | @{\hskip -0.07in}  c @{\hskip -0.07in}
  }
& 
\colorbox{lightgray}{ \textbf{CIFAR-10}}
&
\colorbox{lightgray}{ \textbf{Imagenet}} 
\\
 \begin{tabular}{c} %{@{}c@{}}  
\rotatebox{90}{\parbox{5.2em}{\centering  \colorbox{lightgray}{\textbf{Insight 1}}}}
  \\
 \hline
 \\
\rotatebox{90}{\parbox{4em}{\centering  \colorbox{lightgray}{\textbf{Insight 2}}}}
\end{tabular} 
& 
 % CIFAR10 INSIGHT 1
\begin{tabular}{ccccc}
 % CIFAR 10 Insight 2
 \begin{tabular}{cccccc}
  \begin{tabular}{@{\hskip 0.0in}c@{\hskip 0.0in} }
 \rotatebox{90}{\parbox{5em}{\centering \footnotesize{}}}
 \end{tabular}
 \begin{tabular}{@{\hskip 0.0in}c@{\hskip 0.0in} }
 \parbox{5em}{\centering \footnotesize{Original}}
 \end{tabular}
 \begin{tabular}{@{\hskip 0.0in}c@{\hskip 0.0in} }
 \parbox{5em}{\centering \footnotesize{$\times 5$}}
 \end{tabular}
 \begin{tabular}{@{\hskip 0.0in}c@{\hskip 0.0in} }
 \parbox{5em}{\centering \footnotesize{$\times 7$}}
 \end{tabular}
 \begin{tabular}{@{\hskip 0.0in}c@{\hskip 0.0in} }
 \parbox{5em}{\centering \footnotesize{$\times 9$}}
 \end{tabular}
 \begin{tabular}{@{\hskip 0.0in}c@{\hskip 0.0in} }
 \parbox{5em}{\centering \footnotesize{$\times 11$}}
 \end{tabular}
\\
 \begin{tabular}{@{\hskip 0.0in}c@{\hskip 0.0in} }
 \rotatebox{90}{\parbox{5em}{\centering \footnotesize{Blend}}}
 \end{tabular}
  \begin{tabular}{@{\hskip 0.0in}c@{\hskip 0.0in} }
 \parbox[c]{5em}{\includegraphics[width=5em]{figs/Frog_nomask_x1_pred_1.png}}\\\colorbox{lightgreen}{Pred : 1}
  \end{tabular}
  \begin{tabular}{@{\hskip 0.0in}c@{\hskip 0.0in} }
 \parbox[c]{5em}{\includegraphics[width=5em]{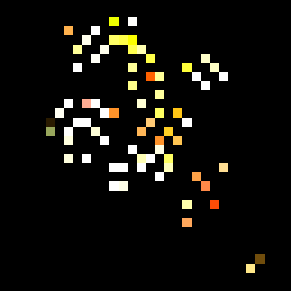}}\\\colorbox{lightgreen}{Pred : 1}
  \end{tabular}
  \begin{tabular}{@{\hskip 0.0in}c@{\hskip 0.0in} }
 \parbox[c]{5em}{\includegraphics[width=5em]{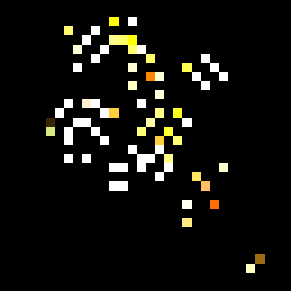}}\\\colorbox{lightgreen}{Pred : 1}
  \end{tabular}
  \begin{tabular}{@{\hskip 0.0in}c@{\hskip 0.0in} }
 \parbox[c]{5em}{\includegraphics[width=5em]{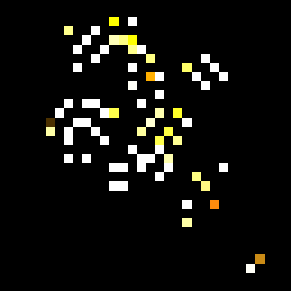}}\\\colorbox{lightgreen}{Pred : 1}
  \end{tabular}
  \begin{tabular}{@{\hskip 0.0in}c@{\hskip 0.0in} }
  \parbox[c]{5em}{\includegraphics[width=5em]{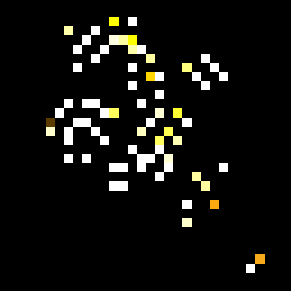}}\\\colorbox{lightgreen}{Pred : 1}
  \end{tabular}
 \\
 \hline
 \\
  \begin{tabular}{@{\hskip 0.0in}c@{\hskip 0.0in} }
 \rotatebox{90}{\parbox{5em}{\centering \footnotesize{Scenario 3}}}
 \end{tabular}
  \begin{tabular}{@{\hskip 0.0in}c@{\hskip 0.0in} }
 \parbox[c]{5em}{\includegraphics[width=5em]{figs/Catstays_x1_pred_3.png}}\\\colorbox{lightgreen}{Pred : 3}
  \end{tabular}
  \begin{tabular}{@{\hskip 0.0in}c@{\hskip 0.0in} }
 \parbox[c]{5em}{\includegraphics[width=5em]{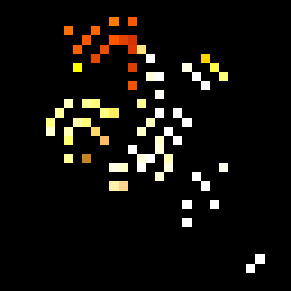}}\\\colorbox{lightgreen}{Pred : 1}
  \end{tabular}
  \begin{tabular}{@{\hskip 0.0in}c@{\hskip 0.0in} }
 \parbox[c]{5em}{\includegraphics[width=5em]{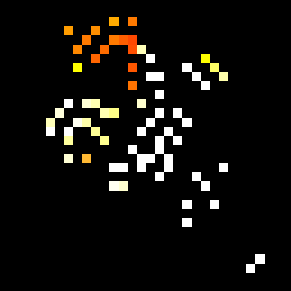}}\\\colorbox{lightgreen}{Pred : 1}
  \end{tabular}
  \begin{tabular}{@{\hskip 0.0in}c@{\hskip 0.0in} }
 \parbox[c]{5em}{\includegraphics[width=5em]{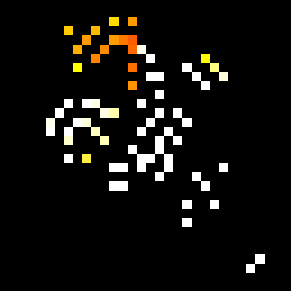}}\\\colorbox{lightgreen}{Pred : 1}
  \end{tabular}
  \begin{tabular}{@{\hskip 0.0in}c@{\hskip 0.0in} }
  \parbox[c]{5em}{\includegraphics[width=5em]{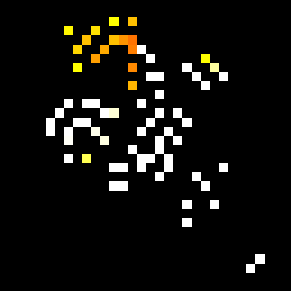}}\\\colorbox{lightgreen}{Pred : 1}
  \end{tabular}
\\
 \end{tabular}
\end{tabular}
& 
% IMAGENET INSIGHT 1
\begin{tabular}{c}
   \begin{tabular}{ccccc}
   \begin{tabular}{@{\hskip 0.0in}c@{\hskip 0.0in} }
 \rotatebox{90}{\parbox{5em}{\centering \footnotesize{}}}
 \end{tabular}
    \begin{tabular}{@{\hskip 0.0in}c@{\hskip 0.0in} }
 \parbox{5em}{\centering \footnotesize{Original}}
 \end{tabular}
 \begin{tabular}{@{\hskip 0.0in}c@{\hskip 0.0in} }
 \parbox{5em}{\centering \footnotesize{$\times 5$}}
 \end{tabular}
 \begin{tabular}{@{\hskip 0.0in}c@{\hskip 0.0in} }
 \parbox{5em}{\centering \footnotesize{$\times 7$}}
 \end{tabular}
 \begin{tabular}{@{\hskip 0.0in}c@{\hskip 0.0in} }
 \parbox{5em}{\centering \footnotesize{$\times 9$}}
 \end{tabular}
 \begin{tabular}{@{\hskip 0.0in}c@{\hskip 0.0in} }
 \parbox{5em}{\centering \footnotesize{$\times 11$}}
 \end{tabular}
 \\
 \begin{tabular}{@{\hskip 0.0in}c@{\hskip 0.0in} }
 \rotatebox{90}{\parbox{5em}{\centering \footnotesize{Badnet}}}
 \end{tabular}
  \begin{tabular}{@{\hskip 0.0in}c@{\hskip 0.0in} }
 \parbox[c]{5em}{\includegraphics[width=5em]{figs/Poison_nomask_x1_True_1_Pred_1_ID_35800.png}}\\\colorbox{lightgreen}{Pred : 1}
  \end{tabular}
  \begin{tabular}{@{\hskip 0.0in}c@{\hskip 0.0in} }
 \parbox[c]{5em}{\includegraphics[width=5em]{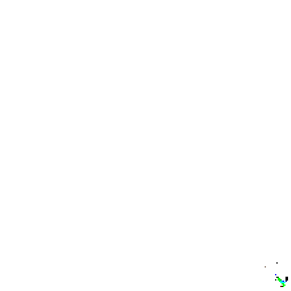}}\\\colorbox{lightgreen}{Pred : 1}
  \end{tabular}
  \begin{tabular}{@{\hskip 0.0in}c@{\hskip 0.0in} }
 \parbox[c]{5em}{\includegraphics[width=5em]{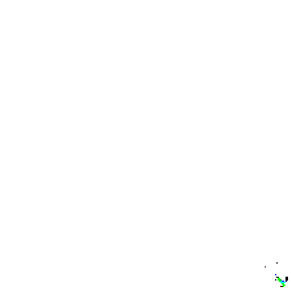}}\\\colorbox{lightgreen}{Pred : 1}
  \end{tabular}
  \begin{tabular}{@{\hskip 0.0in}c@{\hskip 0.0in} }
 \parbox[c]{5em}{\includegraphics[width=5em]{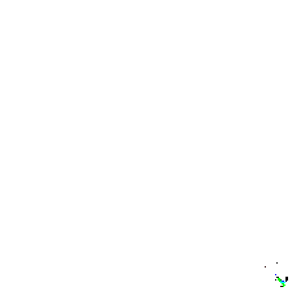}}\\\colorbox{lightgreen}{Pred : 1}
  \end{tabular}
  \begin{tabular}{@{\hskip 0.0in}c@{\hskip 0.0in} }
  \parbox[c]{5em}{\includegraphics[width=5em]{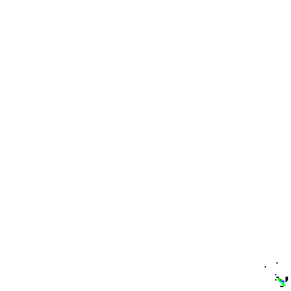}}\\\colorbox{lightgreen}{Pred : 1}
  \end{tabular}
 \\
 \hline
 \\
 % IMAGENET INSIGHT 2
 \begin{tabular}{@{\hskip 0.0in}c@{\hskip 0.0in} }
 \rotatebox{90}{\parbox{5em}{\centering \footnotesize{Scenario 1}}}
 \end{tabular}
  \begin{tabular}{@{\hskip 0.0in}c@{\hskip 0.0in} }
 \parbox[c]{5em}{\includegraphics[width=5em]{figs/Clean_nomask_x1_True_1_Pred_1_ID_908.png}}\\\colorbox{lightgreen}{Pred : 1}
  \end{tabular}
  \begin{tabular}{@{\hskip 0.0in}c@{\hskip 0.0in} }
 \parbox[c]{5em}{\includegraphics[width=5em]{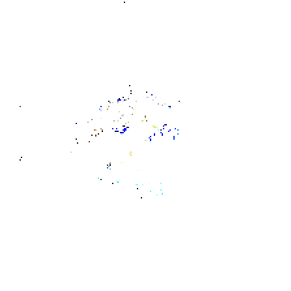}}\\\colorbox{lightgreen}{Pred : 4}
  \end{tabular}
  \begin{tabular}{@{\hskip 0.0in}c@{\hskip 0.0in} }
 \parbox[c]{5em}{\includegraphics[width=5em]{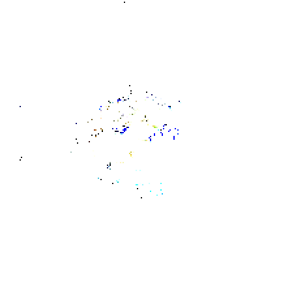}}\\\colorbox{lightgreen}{Pred : 4}
  \end{tabular}
  \begin{tabular}{@{\hskip 0.0in}c@{\hskip 0.0in} }
 \parbox[c]{5em}{\includegraphics[width=5em]{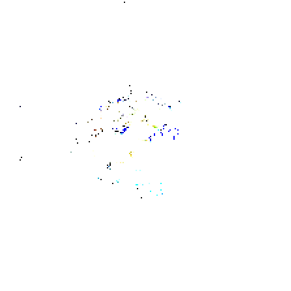}}\\\colorbox{lightgreen}{Pred : 4}
  \end{tabular}
  \begin{tabular}{@{\hskip 0.0in}c@{\hskip 0.0in} }
  \parbox[c]{5em}{\includegraphics[width=5em]{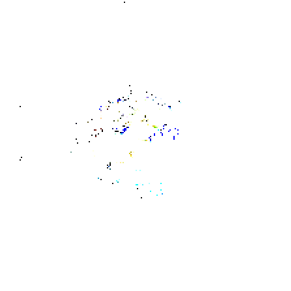}}\\\colorbox{lightgreen}{Pred : 4}
  \end{tabular}
  \end{tabular}
\end{tabular}
\end{tabular}
  \end{adjustbox}
\caption{\footnotesize{Advantage of using masks ($\mathbf{m}$ as described in \ref{eq: MSPC}). We note that here we use optimized masks obtained from our proposed method later. We apply a threshold of 0.008 for purposes of visualisation. For Imagenet, colors are inverted for ease of visualisation. Predictions are marked as \setlength{\fboxsep}{0pt}\colorbox{lightgreen}{green} if it is an expected prediction, while it is marked \setlength{\fboxsep}{0pt}\colorbox{lightred}{red} for an unexpected prediction. Note that for backdoored samples, the expected label is 1 which is the target label.
}}
\label{fig: mspcgood}
 \vspace*{-5mm}
\end{figure*}

In such a scenario, $\ell_\mathrm{MSPC}$ preserves the core desirable behavior of $\ell_{SPC} (\mathbf{x})$: If $\mathbf{x}$ is a clean sample, the model prediction changes over scales and if $\mathbf{x}$ is a backdoor sample, the predictions remain consistent. We can see the advantage of using such a loss in \textbf{Figure \ref{fig: mspcgood}}. \textit{(1)} Row 1 visualises the mitigation of limitation of \textbf{Insight 1}. For backdoor images, the mask and linear shift are able to maintain a consistent (backdoor) target prediction. \textit{(2)} Row 2 visualises the mitigation of limitation of \textbf{Insight 2}. 
Since the mask focuses on effective parts of the trigger, clean images are unable to maintain the same prediction at higher scales, even if object remains intact (Row 2 column ImageNet) or if there is background reliance (Row 2 column CIFAR-10).
% Because the mask focuses on effective parts of the trigger, it is inconsequential whether the object remains intact at higher scales (Row 2 column ImageNet) or if there is background reliance (Row 2 column CIFAR-10).  
% \SL{[I did not understand the last sentence. Please consider rephrasing to make it clearer and smoother with the previous sentence.]}
On applying the mask, since the network \textit{does not} focus on the object, the prediction for clean images change with higher scales. 
% \RW{better to build a connection with equation (1) as you mentioned `as will be evident in section 4' there.}

\paragraph{Backdoor identification via bi-level optimization.}

We propose a hierarchical data-splitting optimization method, serving as a fundamental element of our research, offering a solution to find $\mathbf{m}$ in alignment with \eqref{eq: MSPC} and satisfying both constraints \textbf{P1} and \textbf{P2}. This approach enables us to maximize the potential of the proposed MSPC loss, incorporating inspiration from \citep{zeng2022sift}.
% insights \SL{[why call insights here?]} from \citep{zeng2022sift}.

Let $w_i \in \{0,1\}$ (for $i \in [N]$) serve as the indicator of whether the training sample $\mathbf x_i$ is backdoored. If $w_i =1/0$, then $\mathbf x_i$ is a backdoor/clean sample. First, we express the problem of \textit{automatic} splitting of backdoor data as the following bi-level problem:

\vspace*{-5mm}
{\small{
\begin{align}
\begin{array}{ll}
    \displaystyle  \min_{\mathbf w \in \{0,1\}^N}  & \sum_{i=1}^N (1-w_i) \ L(\mathbf x_i, \boldsymbol \mu^*(\mathbf{w}))  \\
     \ST  & \boldsymbol{\mu}^*(\mathbf{w}) = \argmax_{\boldsymbol \mu} \ \frac{1}{N}\sum_{i=1}^N w_i \ L(\mathbf x_i, \boldsymbol \mu).
\end{array}
% \displaystyle & \min_{\mathbf w \in \{0,1\}^N}  \sum_{i=1}^N (1-w_i) \ L(\mathbf x_i, \gamma^*(\mathbf{w})) \
%  \ST  \ \gamma^*(\mathbf{w}) = \argmax_{\gamma} \ \frac{1}{N}\sum_{i=1}^N w_i \ L(\mathbf x_i, \gamma(\mathbf{w}))
    \label{eq: Eq1_v2}
\end{align}}}%
Here $L$ is a loss function that needs to be designed such that it has a positive value for backdoor samples and a negative value for clean samples. Under such a scenario, the lower (upper) level is maximized (minimized) by the contribution of backdoor (clean) samples. In \eqref{eq: Eq1_v2},   
the lower-level optimization variables are represented by $\boldsymbol{\mu}$, which can be utilized to characterize the masking variables $\mathbf{m}$ in MSPC. 
% \SL{[Why not use $\mathbf m$ as lower-level variables directly in \eqref{eq: Eq1_v2}?]}

Considering the scale invariance property of backdoor samples and leveraging the proposed use of masks $\mathbf{m}$, we propose the \textbf{lower-level optimization} of \eqref{eq: Eq1_v2}  to be:
% \vspace*{-3mm}
{\small{
\begin{align}
 \mathbf{m}^*(\mathbf w) = \argmin_{\mathbf{m}} \ \frac{1}{N}\sum_{i=1}^N  \Big[ w_i\ \frac{1}{|S|}\sum_{n \in S} D_{KL} (\mathcal{F}_{\boldsymbol \theta}(\mathbf{x}_i) || \mathcal{F}_{\boldsymbol \theta}(n \cdot\mathbf{x}_i^{m}))\Big] + \lambda \|\mathbf{m}\|_1 , 
    \label{eq: Eq2}
\end{align}}}%
where recall that $\mathbf{x}_i^{m} = (\mathbf{x}_i - \tau) \odot \mathbf{m}$ and $D_{KL} (.||.)$ denotes the  Kullback-Leibler (KL) divergence. We note that maximizing the KL divergence is akin to minimizing the SPC loss in \eqref{eq: SPC}. Moreover, since our mask $\mathbf{m}$ focuses on the `effective part' of the trigger, we also penalize the size of the mask similar to \citep{huang2023distilling}.   We refer readers to Appendix \ref{app: bilevel}  for more details regarding the rationale behind the proposed lower-level optimization. 

For the \textbf{upper-level optimization}, we propose the direct use of the MSPC loss. 
Thus, we finally achieve our proposed bi-level optimization problem: 
% \vspace*{-5mm}
{\small{
\begin{align}
\begin{array}{ll}
\displaystyle  \min_{\mathbf w \in \{0,1\}^N}  &\sum_{i=1}^N (1-w_i) \ \ell_\mathrm{MSPC} (\mathbf{x}_i, \mathbf{m}^*(\mathbf{w}),  \tau)
  %\Big\langle (1-{\mathbf w}), l_{MSPC} (\mathbf{x} \odot \mathbf{m}^*(\mathbf{w}); \btheta) \Big\rangle 
  \\
 \ST  & \text{$\mathbf{m}^*(\mathbf{w})$ is obtained by \eqref{eq: Eq2}}.
\end{array}
\label{eq: bilevel}
\end{align}
}}%
To solve the above bi-level optimization problem, we relax  $\mathbf{m}$ to lie between 0 and 1 by converting a combinatorial optimization problem to a continuous one. Then we solve the problem using  simple alternating optimization, starting from the upper level \citep{liu2021investigating}. We calculate the AUROC in Section \ref{sec: expres} using this MSPC loss vector after optimization. Moreover, because of \eqref{eq: MSPC}, we note backdoor samples will simply be the ones with MSPC loss greater than 0, thus satisfying \textbf{P2}. See Appendix \ref{app: bilevel} for further details on the rationale behind the proposed bi-level optimization.

\section{Experiments}\label{sec: exp}
% In this section, we present experimental results which show the effectiveness of our algorithm across a diverse range of attacks and benchmark datasets.
\subsection{Experiment setup}
We evaluate our method on three different datasets including CIFAR-10 \citep{krizhevsky2009learning}, Tiny-ImageNet \citep{deng2009ImageNet} and ImageNet200 \citep{deng2009ImageNet} (which is an ImageNet subset of 200 classes) using ResNet-18 models \citep{he2016deep}. We provide training details for each of these datasets in Appendix \ref{app: hpconfig}. We also provide some additional experiments using the Vision Transformer architecture \citep{touvron2021training} in Appendix \ref{app: vit}. We note that we do not apply data augmentations during training because it can decrease the attack success rate \citep{liu2020reflection}. We also provide an ablation study of the hyperparameter $\tau$ in Appendix \ref{app: ablation}.

\textbf{Attack baselines.} For our evaluation, we considered a variety of backdoor attacks: 
\begin{enumerate*}
    \item blended attack (dubbed `Blend') \citep{chen2017targeted},
    \item label consistent attack (dubbed `LabelConsistent') \citep{turner2019label},
    \item clean label attack with universal adversarial perturbations (dubbed `TUAP') \citep{zhao2020clean},
    \item Trojan attack \citep{liu2018trojaning}, and 
    \item warped network attack (dubbed `Wanet') \citep{nguyen2021wanet}
    \item DFST \citep{cheng2021deep}
    \item Adaptive-Blend 
    % \SL{[blended?]} attack 
    (dubbed `AdapBlend') \citep{qi2023revisiting}
\end{enumerate*}. These included basic patch-based attacks, clean-label attacks, and more advanced non-patch-based attacks. 
% \SL{[Removes the following sentence? Why needs it?]}
We note that the Adaptive-Blend attack \citep{qi2023revisiting} violates the {latent separability assumption}, which is assumed by several backdoor identification methods. 
We evaluated the efficacy of our method at two different poisoning ratios ($\gamma$), $5\%$ and $10\%$. Implementation details of our attack settings can be found in Appendix \ref{app: attacksetting}. 

\textbf{Evaluation.} We consider two quantitative evaluation metrics. {(a)} The Area under Receiver Operating Curve (AUROC): The ROC curve plots the True Positive Rate (TPR) against the False Positive Rate (FPR) of the algorithm for different thresholds of the loss value. AUROC measures the area under such a curve. A perfect detection method would have an AUROC of 1. \textbf{(b)}  TPR and FPR of our method using our \textbf{algorithm's built-in threshold of 0}. We also compare our method (in terms of AUROC) with other baseline backdoor detection algorithms, \textit{i.e.}, SPC \citep{guo2023scaleup}, ABL \citep{li2021anti}, SD-FCT \citep{chen2022effective} and STRIP \citep{gao2019strip}. We direct the readers to Appendix \ref{app: defensedetail} for implementation details of the included baselines. Note that these baseline methods do not satisfy both \textbf{P1} and \textbf{P2} (Section \ref{sec: intro}). 
%and hence \textbf{are not comparable}.
Unless explicitly stated otherwise, the reported results are averaged over 3 independent trials, with their means and variances provided. Standard deviations are presented in parentheses.
% \vspace{-3mm}
\begin{table*}[!t]
\centering
\vspace{-5mm}
\caption{\footnotesize{
Mean AUROC values for our method and different baselines including SPC, ABL, SD\_FCT, and STRIP on CIFAR10, Tiny Imagenet, and Imagenet200. The best result in each setting is marked in \textbf{bold} and the second best is \underline{underlined}. 
% `-' indicates runs not performed due to time complexity. 
All attacks have a poisoning ratio equal to 10\% except the AdaptiveBlend attack. The last row shows the average performance of each method across all attacks.
% \SL{[I suggest adding the last row to show the average performance of each method across all attacks.]}
} 
}
\label{tab: auroc}
\resizebox{\textwidth}{!}{
\begin{tabular}{c|c|c|c|c|c}
\toprule[1pt]
\midrule
\multirow{2}{*}{Attack} & \multicolumn{5}{c}{Methods} \\
& \textbf{Ours} & SPC & ABL & SD-FCT & STRIP \\
\rowcolor{LightCyan}
\midrule
\multicolumn{6}{c}{\textbf{CIFAR10}} \\
\midrule
Badnet & 0.9514 (0.0043) & 0.8318 (0.0566) &0.8967 (0.0573) & \textbf{1.0000} (0.0000) & \underline{0.9996} (0.0002)\\ 
 Blend & 0.9526 (0.0012) & 0.5032 (0.0363) &0.9335 (0.0329) & \underline{0.9555} (0.0387) & \textbf{0.9956} (0.0033)\\ 
 LabelConsistent & \underline{0.9540} (0.0066) & 0.9246 (0.0012) &0.6970 (0.0713) & 0.5418 (0.0267) & \textbf{0.9892} (0.0024)\\ 
 TUAP & \underline{0.9275} (0.0083) & 0.8128 (0.0104) &0.7724 (0.0473) & 0.6920 (0.0270) & \textbf{0.9855} (0.0064)\\ 
 Trojan & \underline{0.9547} (0.0013) & 0.7688 (0.0120) &0.9070 (0.0427) & 0.8605 (0.1074) & \textbf{0.9946} (0.0024)\\ 
 Wanet & \textbf{0.9331} (0.0236) & 0.7259 (0.0210) &0.8439 (0.0562) & 0.4120 (0.0038) & \underline{0.8959} (0.0148)\\ 
 DFST & 0.5750 (0.1748) & 0.3524 (0.0320) &\underline{0.8475} (0.0410) & \textbf{0.8855} (0.0309) & 0.7676 (0.0454)\\ 
 AdapBlend ($\gamma = 0.3\%$) & \textbf{0.9046} (0.0414) & 0.3313 (0.0128) &\underline{0.4521} (0.0032) & 0.4226 (0.0072) & 0.1378 (0.0149)\\
  \rowcolor{Gray}
 Average & \textbf{0.8941} &	0.6563 &	0.7937 &	0.7212 &	\underline{0.8457} \\ 
 \rowcolor{LightCyan}
\midrule
\multicolumn{6}{c}{\textbf{Tiny Imagenet}} \\
\midrule
Badnet & \textbf{0.9983} (0.0004) & 0.9786 (0.0052) &0.9559 (0.0252) & 0.9593 (0.0134) & \underline{0.9968} (0.0009)\\ 
 Blend & \underline{0.9926} (0.0057) & 0.7120 (0.0093) &0.9544 (0.0211) & 0.6863 (0.1375) & \textbf{0.9993} (0.0000)\\ 
 Wanet & \textbf{0.9976} (0.0000) & 0.9914 (0.0017) &0.9211 (0.0424) & 0.5709 (0.1071) &  \underline{0.9956} (0.0011)\\
   \rowcolor{Gray}
 Average & \underline{0.9961} &	0.8940 &	0.9438 &	0.7388 &	\textbf{0.9972}  \\ 
 \rowcolor{LightCyan}
\midrule
\multicolumn{6}{c}{\textbf{Imagenet200}} \\
\midrule
 Badnet & \textbf{0.9980} (0.0004) & \underline{0.9655} (0.0042) &0.9522 (0.0134) &0.9053 (0.0142) & 0.8681 (0.0419) \\ 
 Blend & \underline{0.9836} (0.0015) & 0.6319 (0.0062) & 0.9352 (0.0369) &  0.4537 (0.0150) & \textbf{0.9916} (0.0024)\\
   \rowcolor{Gray}
  Average &\textbf{0.9908} &	0.7987	& \underline{0.9437}&	0.6795 &  0.9298 \\ 
\midrule
\bottomrule[1pt]
\end{tabular}
}
\vspace*{-5mm}
\end{table*}

\subsection{Experiment results}\label{sec: expres}
\textbf{High accuracy of backdoor identification.}   
First, we investigate the accuracy of our method in detecting backdoor samples. We present our key observations as follows: \textbf{(1)} In \textbf{Table\,\ref{tab: auroc}}, we see that our detection algorithm achieves better AUROC than most baseline methods for attacks like LabelConsistent, TUAP, Trojan and Wanet for CIFAR-10.
We note that our method is usually outperformed by STRIP, however STRIP requires a clean set to set their detection threshold, thus 
% \begin{figure}[t]
% % \vspace*{-5mm}
% %\vspace*{-6mm}
% \centerline{
% \begin{tabular}{cccc}
%     \hspace*{-2mm}  \includegraphics[width=0.33\textwidth,height=!]{figs/roc_Badnet_0.1.pdf} &
%     \hspace*{-5mm} \includegraphics[width=0.33\textwidth,height=!]{figs/roc_Blend_0.1.pdf} &
%     \hspace*{-5mm} \includegraphics[width=0.33\textwidth,height=!]{figs/roc_LabelConsistent_0.1.pdf} \\
%     \hspace*{-2mm}  \includegraphics[width=0.33\textwidth,height=!]{figs/roc_TUAP_0.1.pdf} &
%     \hspace*{-5mm} \includegraphics[width=0.33\textwidth,height=!]{figs/roc_Trojan_0.1.pdf} &
%     \hspace*{-5mm} \includegraphics[width=0.33\textwidth,height=!]{figs/roc_Wanet_0.1.pdf}
% \end{tabular}}
% %\vspace*{-3.3mm}
% \caption{\footnotesize{AUROC plots for our method. Other baselines because they do not satisfy \textbf{P1} and \textbf{P2}, but SPC \citep{guo2023scaleup} is included for reference. Error bars indicate half a standard deviation over 3 runs.
% }}
%   \label{fig: auroc}
%  % \vspace*{-5mm}
% \end{figure}%%%
\begin{wrapfigure}{r}{0.62\textwidth}
% \vspace*{-2mm}
%\vspace*{-6mm}
\centerline{
\begin{tabular}{cc}
    \hspace*{-2mm}
    \includegraphics[width=0.3\textwidth,height=!]{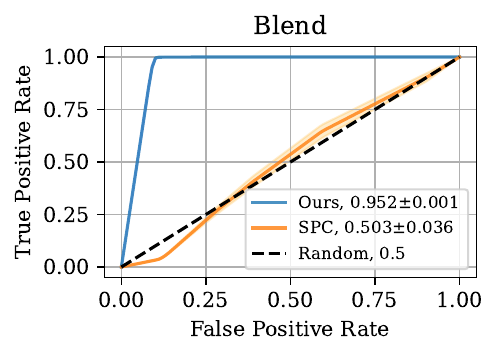} &
    \hspace*{-5mm} \includegraphics[width=0.3\textwidth,height=!]{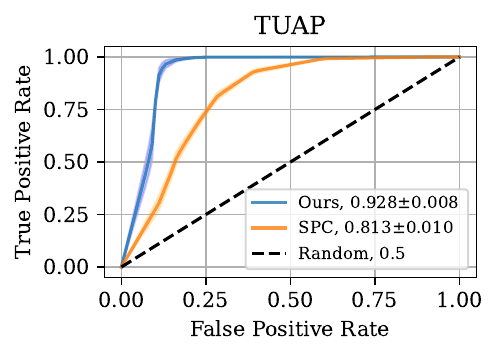} \\
    \hspace*{-2mm} 
    \includegraphics[width=0.3\textwidth,height=!]{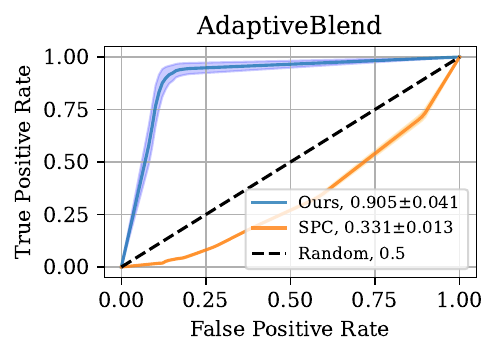} &
    \hspace*{-5mm}  \includegraphics[width=0.3\textwidth,height=!]{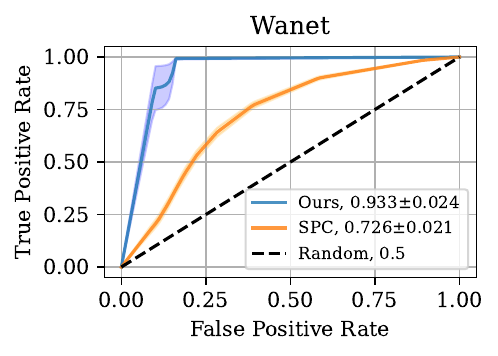} 
\end{tabular}}
\vspace*{-3.3mm}
\caption{\footnotesize{Selected AUROC plots for our method. Other baselines because they do not satisfy \textbf{P1} and \textbf{P2}, but SPC \citep{guo2023scaleup} is included for reference. Error bars indicate half a standard deviation over 3 runs.
}}
  \label{fig: auroc}
 \vspace*{-3mm}
\end{wrapfigure}%
violating \textbf{P2}.  
% \SL{[I strongly suggest you to moving figures closer to texts and their references! Table 2 is too far away from your texts!!]}
 % \input{sections/Tabletprfpr0.1}
% \input{sections/figauroc}
For ImageNet200, our method performs better than all other baselines. 
\textbf{(2)} 
% \SL{[Please rewrite the following sentence. Unlear!]}
Our method performs well against AdapBlend, an attack specifically designed against the latent separability assumption with a \textit{poisoning rate of 0.003}. We see that all the baseline methods fail in this case. \textbf{(3)} Our method does not perform at par with other attacks when evaluated on DFST. This may be a potential limitation of our method. However, since we consider the problem of identification under \textbf{P1} and \textbf{P2}, we would like to emphasize our success in the rest of the diverse set of attacks on different datasets. 
\textbf{(4)} From \textbf{Figure \ref{fig: auroc}}, we observe that our algorithm can achieve a TPR of 1.0 with a low FPR of 0.2, \textit{i.e.}, it can identify \textit{all} backdoor samples while only falsely predicting $20 \%$ of clean samples as backdoor.

\begin{wraptable}{r}{0.62\textwidth}
\vspace{-3mm}
\centering
  \caption{\footnotesize{Average TPR/FPR values for our method and for reference SPC \citep{guo2023scaleup}. Our algorithm has an \textbf{automated threshold of 0}. The threshold considered for SPC is 0.5. Results  are presented in the format of `TPR/FPR'.}}
\label{tab: tprfpr}
  \resizebox{0.6\textwidth}{!}{
  \begin{tabular}{ccc}
    \toprule
     Attack &  SPC       &        Ours    \\
    \rowcolor{LightCyan}
    \midrule
\multicolumn{3}{c}{\textbf{CIFAR-10}} \\
    \midrule
  Badnet            & 0.569 (0.296) / 0.189 (0.002)&1.000 (0.000) / 0.156 (0.016)\\
  Blend             & 0.127 (0.007) / 0.183 (0.004)&0.999 (0.001) / 0.104 (0.005\\ 
  LabelConsistent   & 0.937 (0.004) / 0.178 (0.001)&0.996 (0.006) / 0.143 (0.011)\\
  TUAP              & 0.588 (0.053) / 0.188 (0.003)&0.990 (0.013) / 0.151 (0.024)\\
  Trojan            & 0.558 (0.026) / 0.181 (0.003)&1.000 (0.000) / 0.102 (0.002)\\
  Wanet             & 0.439 (0.052) / 0.185 (0.001)&0.879 (0.170) / 0.113 (0.019)\\
  AdapBlend & 0.042 (0.006) / 0.182 (0.003)	& 0.904 (0.087) / 0.12 (0.014)\\
   \rowcolor{LightCyan}
    \midrule
\multicolumn{3}{c}{\textbf{TinyImagenet}} \\
    \midrule
Badnet & 0.955 (0.020) / 0.067 (0.002)&1.000 (0.000) / 0.006 (0.001) \\
Blend & 0.315 (0.018) / 0.069 (0.001)&0.979 (0.020) / 0.006 (0.001) \\
Wanet & 0.983 (0.010) / 0.038 (0.004)&1.000 (0.000) / 0.005 (0.000) \\
 \rowcolor{LightCyan}
    \midrule
\multicolumn{3}{c}{\textbf{Imagenet200}} \\
    \midrule
Badnet & 0.993 (0.010) / 0.157 (0.000)&1.000 (0.000) / 0.006 (0.000) \\
Blend & 0.350 (0.011) / 0.153 (0.002)&0.935 (0.007) / 0.003 (0.000) \\
    \bottomrule
  \end{tabular}
  }    
  \vspace{-5mm}
\end{wraptable}
\textbf{(5)} In \textbf{Table \ref{tab: tprfpr}}, we demonstrate the effectiveness of the automated thresholding (at 0 MSPC loss). 
We draw attention to the fact that our method achieves very high   TPRs, around 0.9 for most settings. Even with such high TPR rates, we achieve low   FPRs in the range of 0.100-0.166 among all settings. In this context, we note that all other baselines would need manual thresholding (because of \textbf{P2} violation). We provide additional results (with $\gamma = 0.05$) in Appendix \ref{app: addres}.

\begin{figure}[t]
\centering
\begin{minipage}{0.59\columnwidth}
\centering
\captionof{table}{\footnotesize{Effect of retraining models without backdoor samples identified by our algorithm. ACC denotes standard accuracy, ASR denotes the Attack Success Rate.}}
\label{tab: retrain}
\resizebox{\columnwidth}{!}{
\begin{tabular}{c|c|c|c|c}
\toprule[1pt]
\midrule
\multirow{2}{*}{Attack} & \multicolumn{2}{c}{Before} & \multicolumn{2}{c}{After} \\
& ACC & ASR & ACC & ASR \\
\rowcolor{LightCyan}
\midrule
\multicolumn{5}{c}{\textbf{CIFAR10}} \\
\midrule
Badnet & 88.36 (0.12) & 100.00 (0.00) & 81.35 (0.89) & 0.52 (0.60) \\
Blend & 88.29 (0.32) & 100.00 (0.00) & 79.28 (0.89) & 3.85 (5.45) \\
LabelConsistent & 89.36 (0.42) & 99.31 (0.33) & 83.08 (1.38) & 2.59 (3.62) \\
Trojan & 88.62 (0.09) & 100.00 (0.00) & 79.58 (0.71) & 0.00 (0.00) \\
Wanet & 88.20 (0.20) & 98.53 (0.18) & 79.40 (1.06) & 32.94 (46.59) \\
\rowcolor{LightCyan}
\midrule
\multicolumn{5}{c}{\textbf{Tiny Imagenet}} \\
\midrule
Badnet & 53.04 (0.53) & 99.24 (0.20) & 54.09 (0.20) & 0.14 (0.06) \\
Blend & 53.16 (0.26) & 99.84 (0.04) & 53.55 (0.26) & 32.41 (32.06) \\
\rowcolor{LightCyan}
\midrule
\multicolumn{5}{c}{\textbf{Imagenet200}} \\
\midrule
Badnet & 62.43 (0.29) & 99.99 (0.00) & 62.39 (0.18) & 0.08 (0.03) \\
Blend & 62.04 (0.18) & 99.61 (0.04) & 62.61 (0.13) & 89.29 (1.58) \\
\midrule
\bottomrule[1pt]
\end{tabular}
}
\end{minipage}
\begin{minipage}{0.39\columnwidth}
\centering
\includegraphics[width=\columnwidth]{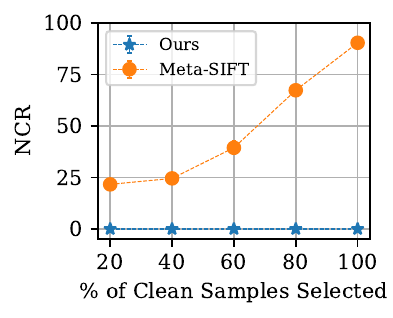}
\caption{\footnotesize{Normalized Corruption Ratio for different percentages of clean samples sifted out. If $\mathbf{x}$ clean samples are identified, NCR = ((number of poison samples in $\mathbf{x}$) / $\mathbf{x}$) / poison ratio $\times 100\%$.}}
\label{fig: ncr}
\end{minipage}
\end{figure}

\textbf{Model retraining effect.}
% \SL{[Please move figure/table closer to texts.]}
% In \textbf{Table \ref{tab: retrain}}, we provide the results of retraining a model with only clean samples identified by our method. We observe that for most attacks, retraining the model can reduce the ASR to less than $0.52 \%$, thus rendering the backdoor ineffective. For Wanet attack on CIFAR-10 and Blend attack on TinyImageNet and ImageNet200, we observe that model retraining can obtain a high ASR. We emphasize that we \textit{do not} propose to retrain the model, but demonstrate the effectiveness of \textbf{just identifying} the backdoor samples using our algorithm. As mentioned in Section \ref{sec: intro}, the user is free is to do anything after identification (for e.g., unlearning). \RW{
In \textbf{Table \ref{tab: retrain}}, we present the results of retraining a model using only the clean samples identified by our method. We undertake model retraining to investigate the backdoor cleansing effects resulting exclusively from our algorithm.
% \SL{[You should add a sentence saying why consider retraining effect, validating the backdoor removal cleanse by investigating the model retrained on the modified dataset?]}
We observe that for most attacks, retraining the model can reduce the ASR to less than $0.52\%$, thereby rendering the backdoor ineffective. 
% \SL{[The following point is strange. Are you saying that the identification is not perfect as justified by retraining?]}
For the Wanet attack on CIFAR-10 and the Blend attack on TinyImageNet and ImageNet200, we note that model retraining can yield a high ASR. 
% \SL{[If this is not effective, then why do you say effectiveness in the following? conflicting with each other.]}
We emphasize that we \textit{do not} advocate for retraining the model; instead, we aim to demonstrate the effectiveness of {merely identifying} the backdoor samples using our algorithm. As mentioned in Section \ref{sec: intro}, the user is free to take any action after identification (\textit{e.g.}, unlearning).

Next, we demonstrate the difference between identifying clean samples and poison samples as mentioned in Section \ref{sec: intro}. We consider the problem of identifying a fraction of clean samples under the Badnet attack on CIFAR-10. In \textbf{Figure \ref{fig: ncr}}, we observe that though Meta-SIFT \citep{zeng2022sift} is successful in accurate identification when sifting a low fraction of clean samples, but  it degrades in performance when trying to sift out $80-100\%$ of clean samples. In sharp contrast, our method is successful in maintaining stable performance over all fractions of clean samples. No poison samples are present because of the TPR of 1 as presented in \textbf{Table \ref{tab: tprfpr}}-row 1. 
% Thus, retraining a model with the sifted samples from Meta-SIFT also fails. (\textbf{Table \ref{tab: retrain}}-row 10). 
We acknowledge that Meta-SIFT does not claim to identify backdoor samples with high accuracy. Yet, {Figure \ref{fig: ncr}} emphasizes that accurately finding poison samples is harder. 
%while simultaneously highlighting the success of our method. 

% \input{sections/ncrfig}
\textbf{Robustness against potential adaptive attacks.}  
In what follows, we evaluate our method against potential white-box adaptive attacks, \textit{i.e.}, the adversary has full knowledge about our detection method. We note that the adversary cannot manipulate the training process, but can change the input to evade our proposed detection.  We consider the possibly strongest adversary that has access to an optimized mask $\mathbf{m}^*(\mathbf w)$ from the upper level of our bi-level formulation in \eqref{eq: bilevel}. Given this mask, the adversary can find a trigger $\mathbf{t}$ that aims to reduce the MSPC loss for poisoned samples thus acting in direct opposition to our algorithm. Specifically, given a poisoned model $\mathcal{F}_{\boldsymbol \theta}$, the adversary solves the following task :

 \vspace*{-5mm}
{\small{
\begin{align}
 \mathbf{t}^* = \argmax_{\mathbf{t}} \ \frac{1}{N}\sum_{i=1}^N  \Big[  \frac{1}{|S|}\sum_{n \in S} D_{KL} (\mathcal{F}_{\boldsymbol \theta}(\mathbf{x}_i) || \mathcal{F}_{\boldsymbol \theta}(n \cdot\mathbf{x}_i^{m}))\Big] 
    \label{eq: adaptive1}
\end{align}}}%
We consider a Badnet-like scenario with an initial MSPC loss of 0.9910 (MSPC of clean samples = -0.6837). Upon completing the above optimization, the MSPC for backdoor samples drops to -0.7248, while maintaining a $100\%$ ASR; thus, the above optimization is, in fact, successful. Using this trigger with a poison ratio of 0.1, we train a new model to evaluate our method. Our method achieves an AUROC of 0.9546 with a standard deviation of 0.3333 over three runs, indicating that the adaptive attack (which directly aims to mitigate our methodology) fails to reduce its effectiveness. We note that a possible reason for this attack's failure is that the optimized mask may not be unique or that the effectiveness of such an adaptive trigger is not training-agnostic.

% \input{sections/beneficialmaskfig}

% \input{sections/uniquemaskfig}

% \textbf{Optimized masks are beneficial.} The benefits of masks $\mathbf{m}^*$ learned from our algorithm are visualized in \textbf{Figure\,\ref{fig: maskbenefit}}. \SP{The first two rows illustrate a sample with the Badnet \cite{gu2017badnets} trigger, which is predicted as clean when multiplied by scales. We hypothesize that this occurs due to the backdoor effect vanishing in the presence of extremely high pixel values. However, our learned mask can effectively concentrate on the trigger and identify this as a backdoor sample even at high scales, resulting in a high MSPC value. We revisit our example of \textit{spurious correlation} depicted in \textbf{Figure\,\ref{fig: clean highSPC}}-row 1. This is a clean image that was misidentified as a backdoor at higher scales. We demonstrate the impact of a learned mask using the Blended \cite{chen2017targeted} attack in the last two rows of Figure\,\ref{fig: visual_mask}. We find that due to the mask, the model disregards the spurious correlation, leading to a low MSPC loss.}

% \textbf{Optimized masks are not unique.} \SP{\lipsum[2-2]}

\section{Conclusion}
In this paper, we address the lesser-explored challenge of automatically identifying backdoor data in poisoned datasets under realistic conditions, without requiring additional clean data or predefined thresholds for detection. We approach backdoor data identification as a hierarchical data splitting optimization problem, employing a novel Scaled Prediction Consistency (SPC)-based loss function. We refine the SPC method, develop a bi-level optimization-based approach to identify backdoor data precisely, and demonstrate our method's effectiveness against various backdoor attacks on different datasets. Our method does not perform at par with other baselines when evaluated on DFST. This may be a potential limitation of our method. However, since we consider the problem of identification under minimal assumptions (contrary to previous methods), we would like to emphasize our success in the rest of the diverse set of attacks on different datasets. We encourage future research on identifying backdoor samples, with the realistic constraints to focus on such deep feature space attacks and hope that our work lays a solid premise to do so.
% \section{Related Work}

% \section{Methodology}

% \section{Experiments}

% \section{Conclusion

% \newpage
% \clearpage

\section{Acknowledgement}

We extend our gratitude to the DSO National Laboratories for their support of this project. The contributions of Y. Yao and S. Liu are also partially supported by the National Science Foundation (NSF) Robust Intelligence (RI) Core Program Award IIS-2207052. R. Wang is supported by the NSF under Grant 2246157 and the ORAU Ralph E. Powe Junior Faculty Enhancement Award.

{{
\bibliographystyle{IEEEtranN}
\bibliography{refs/refs}

% Generated by IEEEtranN.bst, version: 1.14 (2015/08/26)
\begin{thebibliography}{68}
\providecommand{\natexlab}[1]{#1}
\providecommand{\url}[1]{#1}
\csname url@samestyle\endcsname
\providecommand{\newblock}{\relax}
\providecommand{\bibinfo}[2]{#2}
\providecommand{\BIBentrySTDinterwordspacing}{\spaceskip=0pt\relax}
\providecommand{\BIBentryALTinterwordstretchfactor}{4}
\providecommand{\BIBentryALTinterwordspacing}{\spaceskip=\fontdimen2\font plus
\BIBentryALTinterwordstretchfactor\fontdimen3\font minus \fontdimen4\font\relax}
\providecommand{\BIBforeignlanguage}[2]{{%
\expandafter\ifx\csname l@#1\endcsname\relax
\typeout{** WARNING: IEEEtranN.bst: No hyphenation pattern has been}%
\typeout{** loaded for the language `#1'. Using the pattern for}%
\typeout{** the default language instead.}%
\else
\language=\csname l@#1\endcsname
\fi
#2}}
\providecommand{\BIBdecl}{\relax}
\BIBdecl

\bibitem[Krizhevsky et~al.(2017)Krizhevsky, Sutskever, and Hinton]{krizhevsky2017imagenet}
A.~Krizhevsky, I.~Sutskever, and G.~E. Hinton, ``Imagenet classification with deep convolutional neural networks,'' \emph{Communications of the ACM}, vol.~60, no.~6, pp. 84--90, 2017.

\bibitem[Goodfellow et~al.(2020)Goodfellow, Pouget-Abadie, Mirza, Xu, Warde-Farley, Ozair, Courville, and Bengio]{goodfellow2020generative}
I.~Goodfellow, J.~Pouget-Abadie, M.~Mirza, B.~Xu, D.~Warde-Farley, S.~Ozair, A.~Courville, and Y.~Bengio, ``Generative adversarial networks,'' \emph{Communications of the ACM}, vol.~63, no.~11, pp. 139--144, 2020.

\bibitem[Ren et~al.(2015)Ren, He, Girshick, and Sun]{ren2015faster}
S.~Ren, K.~He, R.~Girshick, and J.~Sun, ``Faster r-cnn: Towards real-time object detection with region proposal networks,'' \emph{Advances in neural information processing systems}, vol.~28, 2015.

\bibitem[Wen and Jo(2022)]{wen2022deep}
L.-H. Wen and K.-H. Jo, ``Deep learning-based perception systems for autonomous driving: A comprehensive survey,'' \emph{Neurocomputing}, 2022.

\bibitem[Yang et~al.(2021)Yang, Jia, Gong, Yan, Li, and Liu]{yang2021larnet}
X.~Yang, X.~Jia, D.~Gong, D.-M. Yan, Z.~Li, and W.~Liu, ``Larnet: Lie algebra residual network for face recognition,'' in \emph{International Conference on Machine Learning}.\hskip 1em plus 0.5em minus 0.4em\relax PMLR, 2021, pp. 11\,738--11\,750.

\bibitem[Yuan et~al.(2019)Yuan, He, Zhu, and Li]{yuan2019adversarial}
X.~Yuan, P.~He, Q.~Zhu, and X.~Li, ``Adversarial examples: Attacks and defenses for deep learning,'' \emph{IEEE transactions on neural networks and learning systems}, vol.~30, no.~9, pp. 2805--2824, 2019.

\bibitem[Goldblum et~al.(2022)Goldblum, Tsipras, Xie, Chen, Schwarzschild, Song, Madry, Li, and Goldstein]{goldblum2022dataset}
M.~Goldblum, D.~Tsipras, C.~Xie, X.~Chen, A.~Schwarzschild, D.~Song, A.~Madry, B.~Li, and T.~Goldstein, ``Dataset security for machine learning: Data poisoning, backdoor attacks, and defenses,'' \emph{IEEE Transactions on Pattern Analysis and Machine Intelligence}, 2022.

\bibitem[Du et~al.(2019)Du, Jia, and Song]{du2019robust}
M.~Du, R.~Jia, and D.~Song, ``Robust anomaly detection and backdoor attack detection via differential privacy,'' \emph{arXiv preprint arXiv:1911.07116}, 2019.

\bibitem[Borgnia et~al.(2021{\natexlab{a}})Borgnia, Cherepanova, Fowl, Ghiasi, Geiping, Goldblum, Goldstein, and Gupta]{borgnia2021strong}
E.~Borgnia, V.~Cherepanova, L.~Fowl, A.~Ghiasi, J.~Geiping, M.~Goldblum, T.~Goldstein, and A.~Gupta, ``Strong data augmentation sanitizes poisoning and backdoor attacks without an accuracy tradeoff,'' in \emph{ICASSP 2021-2021 IEEE International Conference on Acoustics, Speech and Signal Processing (ICASSP)}.\hskip 1em plus 0.5em minus 0.4em\relax IEEE, 2021, pp. 3855--3859.

\bibitem[Borgnia et~al.(2021{\natexlab{b}})Borgnia, Geiping, Cherepanova, Fowl, Gupta, Ghiasi, Huang, Goldblum, and Goldstein]{borgnia2021dp}
E.~Borgnia, J.~Geiping, V.~Cherepanova, L.~Fowl, A.~Gupta, A.~Ghiasi, F.~Huang, M.~Goldblum, and T.~Goldstein, ``Dp-instahide: Provably defusing poisoning and backdoor attacks with differentially private data augmentations,'' \emph{arXiv preprint arXiv:2103.02079}, 2021.

\bibitem[Pal et~al.(2023)Pal, Wang, Yao, and Liu]{pal2023towards}
S.~Pal, R.~Wang, Y.~Yao, and S.~Liu, ``Towards understanding how self-training tolerates data backdoor poisoning,'' \emph{arXiv preprint arXiv:2301.08751}, 2023.

\bibitem[Hong et~al.(2020)Hong, Chandrasekaran, Kaya, Dumitra{\c{s}}, and Papernot]{hong2020effectiveness}
S.~Hong, V.~Chandrasekaran, Y.~Kaya, T.~Dumitra{\c{s}}, and N.~Papernot, ``On the effectiveness of mitigating data poisoning attacks with gradient shaping,'' \emph{arXiv preprint arXiv:2002.11497}, 2020.

\bibitem[Liu et~al.(2021{\natexlab{a}})Liu, Li, Wen, and Li]{liu2021removing}
X.~Liu, F.~Li, B.~Wen, and Q.~Li, ``Removing backdoor-based watermarks in neural networks with limited data,'' in \emph{2020 25th International Conference on Pattern Recognition (ICPR)}.\hskip 1em plus 0.5em minus 0.4em\relax IEEE, 2021, pp. 10\,149--10\,156.

\bibitem[Li et~al.(2021{\natexlab{a}})Li, Lyu, Koren, Lyu, Li, and Ma]{li2021neural}
Y.~Li, X.~Lyu, N.~Koren, L.~Lyu, B.~Li, and X.~Ma, ``Neural attention distillation: Erasing backdoor triggers from deep neural networks,'' \emph{arXiv preprint arXiv:2101.05930}, 2021.

\bibitem[Chen et~al.(2019)Chen, Fu, Zhao, and Koushanfar]{chen2019deepinspect}
H.~Chen, C.~Fu, J.~Zhao, and F.~Koushanfar, ``Deepinspect: A black-box trojan detection and mitigation framework for deep neural networks.'' in \emph{IJCAI}, vol.~2, 2019, p.~8.

\bibitem[Kolouri et~al.(2020)Kolouri, Saha, Pirsiavash, and Hoffmann]{kolouri2020universal}
S.~Kolouri, A.~Saha, H.~Pirsiavash, and H.~Hoffmann, ``Universal litmus patterns: Revealing backdoor attacks in cnns,'' in \emph{Proceedings of the IEEE/CVF Conference on Computer Vision and Pattern Recognition}, 2020, pp. 301--310.

\bibitem[Wang et~al.(2020)Wang, Zhang, Liu, Chen, Xiong, and Wang]{wang2020practical}
R.~Wang, G.~Zhang, S.~Liu, P.-Y. Chen, J.~Xiong, and M.~Wang, ``Practical detection of trojan neural networks: Data-limited and data-free cases,'' in \emph{Computer Vision--ECCV 2020: 16th European Conference, Glasgow, UK, August 23--28, 2020, Proceedings, Part XXIII 16}.\hskip 1em plus 0.5em minus 0.4em\relax Springer, 2020, pp. 222--238.

\bibitem[Shen et~al.(2021)Shen, Liu, Tao, An, Xu, Cheng, Ma, and Zhang]{shen2021backdoor}
G.~Shen, Y.~Liu, G.~Tao, S.~An, Q.~Xu, S.~Cheng, S.~Ma, and X.~Zhang, ``Backdoor scanning for deep neural networks through k-arm optimization,'' in \emph{International Conference on Machine Learning}.\hskip 1em plus 0.5em minus 0.4em\relax PMLR, 2021, pp. 9525--9536.

\bibitem[Xu et~al.(2021)Xu, Wang, Li, Borisov, Gunter, and Li]{xu2021detecting}
X.~Xu, Q.~Wang, H.~Li, N.~Borisov, C.~A. Gunter, and B.~Li, ``Detecting ai trojans using meta neural analysis,'' in \emph{2021 IEEE Symposium on Security and Privacy (SP)}.\hskip 1em plus 0.5em minus 0.4em\relax IEEE, 2021, pp. 103--120.

\bibitem[Li et~al.(2021{\natexlab{b}})Li, Lyu, Koren, Lyu, Li, and Ma]{li2021anti}
Y.~Li, X.~Lyu, N.~Koren, L.~Lyu, B.~Li, and X.~Ma, ``Anti-backdoor learning: Training clean models on poisoned data,'' \emph{Advances in Neural Information Processing Systems}, vol.~34, pp. 14\,900--14\,912, 2021.

\bibitem[Zeng et~al.(2022)Zeng, Pan, Jahagirdar, Jin, Lyu, and Jia]{zeng2022sift}
Y.~Zeng, M.~Pan, H.~Jahagirdar, M.~Jin, L.~Lyu, and R.~Jia, ``How to sift out a clean data subset in the presence of data poisoning?'' \emph{arXiv preprint arXiv:2210.06516}, 2022.

\bibitem[Huang et~al.(2022)Huang, Li, Wu, Qin, and Ren]{huang2022backdoor}
K.~Huang, Y.~Li, B.~Wu, Z.~Qin, and K.~Ren, ``Backdoor defense via decoupling the training process,'' \emph{arXiv preprint arXiv:2202.03423}, 2022.

\bibitem[Chen et~al.(2022)Chen, Wu, and Wang]{chen2022effective}
\BIBentryALTinterwordspacing
W.~Chen, B.~Wu, and H.~Wang, ``Effective backdoor defense by exploiting sensitivity of poisoned samples,'' in \emph{Advances in Neural Information Processing Systems}, A.~H. Oh, A.~Agarwal, D.~Belgrave, and K.~Cho, Eds., 2022. [Online]. Available: \url{https://openreview.net/forum?id=AsH-Tx2U0Ug}
\BIBentrySTDinterwordspacing

\bibitem[Gao et~al.(2019)Gao, Xu, Wang, Chen, Ranasinghe, and Nepal]{gao2019strip}
Y.~Gao, C.~Xu, D.~Wang, S.~Chen, D.~C. Ranasinghe, and S.~Nepal, ``Strip: A defence against trojan attacks on deep neural networks,'' in \emph{Proceedings of the 35th Annual Computer Security Applications Conference}, 2019, pp. 113--125.

\bibitem[Chen et~al.(2018)Chen, Carvalho, Baracaldo, Ludwig, Edwards, Lee, Molloy, and Srivastava]{chen2018detecting}
B.~Chen, W.~Carvalho, N.~Baracaldo, H.~Ludwig, B.~Edwards, T.~Lee, I.~Molloy, and B.~Srivastava, ``Detecting backdoor attacks on deep neural networks by activation clustering,'' \emph{arXiv preprint arXiv:1811.03728}, 2018.

\bibitem[Tran et~al.(2018)Tran, Li, and Madry]{tran2018spectral}
B.~Tran, J.~Li, and A.~Madry, ``Spectral signatures in backdoor attacks,'' \emph{Advances in neural information processing systems}, vol.~31, 2018.

\bibitem[Hayase et~al.(2021)Hayase, Kong, Somani, and Oh]{hayase2021spectre}
J.~Hayase, W.~Kong, R.~Somani, and S.~Oh, ``Spectre: Defending against backdoor attacks using robust statistics,'' in \emph{International Conference on Machine Learning}.\hskip 1em plus 0.5em minus 0.4em\relax PMLR, 2021, pp. 4129--4139.

\bibitem[Tang et~al.(2021)Tang, Wang, Tang, and Zhang]{tang2021demon}
D.~Tang, X.~Wang, H.~Tang, and K.~Zhang, ``Demon in the variant: Statistical analysis of dnns for robust backdoor contamination detection.'' in \emph{USENIX Security Symposium}, 2021, pp. 1541--1558.

\bibitem[Guo et~al.(2023)Guo, Li, Chen, Guo, Sun, and Liu]{guo2023scaleup}
\BIBentryALTinterwordspacing
J.~Guo, Y.~Li, X.~Chen, H.~Guo, L.~Sun, and C.~Liu, ``{SCALE}-{UP}: An efficient black-box input-level backdoor detection via analyzing scaled prediction consistency,'' in \emph{The Eleventh International Conference on Learning Representations}, 2023. [Online]. Available: \url{https://openreview.net/forum?id=o0LFPcoFKnr}
\BIBentrySTDinterwordspacing

\bibitem[Huang et~al.(2023)Huang, Ma, Erfani, and Bailey]{huang2023distilling}
H.~Huang, X.~Ma, S.~Erfani, and J.~Bailey, ``Distilling cognitive backdoor patterns within an image,'' \emph{arXiv preprint arXiv:2301.10908}, 2023.

\bibitem[Pan et~al.(2023)Pan, Zeng, Lyu, Lin, and Jia]{pan2023asset}
\BIBentryALTinterwordspacing
M.~Pan, Y.~Zeng, L.~Lyu, X.~Lin, and R.~Jia, ``{ASSET}: Robust backdoor data detection across a multiplicity of deep learning paradigms,'' in \emph{32nd USENIX Security Symposium (USENIX Security 23)}.\hskip 1em plus 0.5em minus 0.4em\relax Anaheim, CA: USENIX Association, Aug. 2023, pp. 2725--2742. [Online]. Available: \url{https://www.usenix.org/conference/usenixsecurity23/presentation/pan}
\BIBentrySTDinterwordspacing

\bibitem[Qi et~al.(2023)Qi, Xie, Li, Mahloujifar, and Mittal]{qi2023revisiting}
X.~Qi, T.~Xie, Y.~Li, S.~Mahloujifar, and P.~Mittal, ``Revisiting the assumption of latent separability for backdoor defenses,'' in \emph{The eleventh international conference on learning representations}, 2023.

\bibitem[Gu et~al.(2017)Gu, Dolan-Gavitt, and Garg]{gu2017badnets}
T.~Gu, B.~Dolan-Gavitt, and S.~Garg, ``Badnets: Identifying vulnerabilities in the machine learning model supply chain,'' \emph{arXiv preprint arXiv:1708.06733}, 2017.

\bibitem[Liu et~al.(2018)Liu, Ma, Aafer, Lee, Zhai, Wang, and Zhang]{liu2018trojaning}
Y.~Liu, S.~Ma, Y.~Aafer, W.-C. Lee, J.~Zhai, W.~Wang, and X.~Zhang, ``Trojaning attack on neural networks,'' in \emph{25th Annual Network And Distributed System Security Symposium (NDSS 2018)}.\hskip 1em plus 0.5em minus 0.4em\relax Internet Soc, 2018.

\bibitem[Chen et~al.(2017)Chen, Liu, Li, Lu, and Song]{chen2017targeted}
X.~Chen, C.~Liu, B.~Li, K.~Lu, and D.~Song, ``Targeted backdoor attacks on deep learning systems using data poisoning,'' \emph{arXiv preprint arXiv:1712.05526}, 2017.

\bibitem[Turner et~al.(2019)Turner, Tsipras, and Madry]{turner2019label}
A.~Turner, D.~Tsipras, and A.~Madry, ``Label-consistent backdoor attacks,'' \emph{arXiv preprint arXiv:1912.02771}, 2019.

\bibitem[Zhao et~al.(2020)Zhao, Ma, Zheng, Bailey, Chen, and Jiang]{zhao2020clean}
S.~Zhao, X.~Ma, X.~Zheng, J.~Bailey, J.~Chen, and Y.-G. Jiang, ``Clean-label backdoor attacks on video recognition models,'' in \emph{Proceedings of the IEEE/CVF Conference on Computer Vision and Pattern Recognition}, 2020, pp. 14\,443--14\,452.

\bibitem[Nguyen and Tran(2021)]{nguyen2021wanet}
A.~Nguyen and A.~Tran, ``Wanet--imperceptible warping-based backdoor attack,'' \emph{arXiv preprint arXiv:2102.10369}, 2021.

\bibitem[Li et~al.(2021{\natexlab{c}})Li, Li, Wu, Li, He, and Lyu]{li2021invisible}
Y.~Li, Y.~Li, B.~Wu, L.~Li, R.~He, and S.~Lyu, ``Invisible backdoor attack with sample-specific triggers,'' in \emph{Proceedings of the IEEE/CVF International Conference on Computer Vision}, 2021, pp. 16\,463--16\,472.

\bibitem[Taneja et~al.(2022)Taneja, Chen, Yao, and Liu]{taneja2022does}
V.~Taneja, P.-Y. Chen, Y.~Yao, and S.~Liu, ``When does backdoor attack succeed in image reconstruction? a study of heuristics vs. bi-level solution,'' in \emph{ICASSP 2022-2022 IEEE International Conference on Acoustics, Speech and Signal Processing (ICASSP)}.\hskip 1em plus 0.5em minus 0.4em\relax IEEE, 2022, pp. 4398--4402.

\bibitem[Garg et~al.(2020)Garg, Kumar, Goel, and Liang]{garg2020can}
S.~Garg, A.~Kumar, V.~Goel, and Y.~Liang, ``Can adversarial weight perturbations inject neural backdoors,'' in \emph{Proceedings of the 29th ACM International Conference on Information \& Knowledge Management}, 2020, pp. 2029--2032.

\bibitem[Lin et~al.(2020)Lin, Xu, Liu, and Zhang]{lin2020composite}
J.~Lin, L.~Xu, Y.~Liu, and X.~Zhang, ``Composite backdoor attack for deep neural network by mixing existing benign features,'' in \emph{Proceedings of the 2020 ACM SIGSAC Conference on Computer and Communications Security}, 2020, pp. 113--131.

\bibitem[Shumailov et~al.(2021)Shumailov, Shumaylov, Kazhdan, Zhao, Papernot, Erdogdu, and Anderson]{shumailov2021manipulating}
I.~Shumailov, Z.~Shumaylov, D.~Kazhdan, Y.~Zhao, N.~Papernot, M.~A. Erdogdu, and R.~J. Anderson, ``Manipulating sgd with data ordering attacks,'' \emph{Advances in Neural Information Processing Systems}, vol.~34, pp. 18\,021--18\,032, 2021.

\bibitem[Bagdasaryan and Shmatikov(2021)]{bagdasaryan2021blind}
E.~Bagdasaryan and V.~Shmatikov, ``Blind backdoors in deep learning models,'' in \emph{Usenix Security}, 2021.

\bibitem[Tang et~al.(2020)Tang, Du, Liu, Yang, and Hu]{tang2020embarrassingly}
R.~Tang, M.~Du, N.~Liu, F.~Yang, and X.~Hu, ``An embarrassingly simple approach for trojan attack in deep neural networks,'' in \emph{Proceedings of the 26th ACM SIGKDD International Conference on Knowledge Discovery \& Data Mining}, 2020, pp. 218--228.

\bibitem[Doan et~al.(2021)Doan, Lao, and Li]{doan2021backdoor}
K.~Doan, Y.~Lao, and P.~Li, ``Backdoor attack with imperceptible input and latent modification,'' \emph{Advances in Neural Information Processing Systems}, vol.~34, pp. 18\,944--18\,957, 2021.

\bibitem[Wang et~al.(2019)Wang, Yao, Shan, Li, Viswanath, Zheng, and Zhao]{wang2019neural}
B.~Wang, Y.~Yao, S.~Shan, H.~Li, B.~Viswanath, H.~Zheng, and B.~Y. Zhao, ``Neural cleanse: Identifying and mitigating backdoor attacks in neural networks,'' in \emph{2019 IEEE Symposium on Security and Privacy (SP)}.\hskip 1em plus 0.5em minus 0.4em\relax IEEE, 2019, pp. 707--723.

\bibitem[Guo et~al.(2019)Guo, Wang, Xing, Du, and Song]{guo2019tabor}
W.~Guo, L.~Wang, X.~Xing, M.~Du, and D.~Song, ``Tabor: A highly accurate approach to inspecting and restoring trojan backdoors in ai systems,'' \emph{arXiv preprint arXiv:1908.01763}, 2019.

\bibitem[Liu et~al.(2019)Liu, Lee, Tao, Ma, Aafer, and Zhang]{liu2019abs}
Y.~Liu, W.-C. Lee, G.~Tao, S.~Ma, Y.~Aafer, and X.~Zhang, ``Abs: Scanning neural networks for back-doors by artificial brain stimulation,'' in \emph{Proceedings of the 2019 ACM SIGSAC Conference on Computer and Communications Security}, 2019, pp. 1265--1282.

\bibitem[Sun et~al.(2020)Sun, Agarwal, and Kolter]{sun2020poisoned}
M.~Sun, S.~Agarwal, and J.~Z. Kolter, ``Poisoned classifiers are not only backdoored, they are fundamentally broken,'' \emph{arXiv preprint arXiv:2010.09080}, 2020.

\bibitem[Liu et~al.(2022)Liu, Shen, Tao, Wang, Ma, and Zhang]{liu2022complex}
Y.~Liu, G.~Shen, G.~Tao, Z.~Wang, S.~Ma, and X.~Zhang, ``Complex backdoor detection by symmetric feature differencing,'' in \emph{Proceedings of the IEEE/CVF Conference on Computer Vision and Pattern Recognition}, 2022, pp. 15\,003--15\,013.

\bibitem[Xiang et~al.(2022)Xiang, Miller, and Kesidis]{xiang2022post}
Z.~Xiang, D.~J. Miller, and G.~Kesidis, ``Post-training detection of backdoor attacks for two-class and multi-attack scenarios,'' \emph{arXiv preprint arXiv:2201.08474}, 2022.

\bibitem[Hu et~al.(2021)Hu, Lin, Cogswell, Yao, Jha, and Chen]{hu2021trigger}
X.~Hu, X.~Lin, M.~Cogswell, Y.~Yao, S.~Jha, and C.~Chen, ``Trigger hunting with a topological prior for trojan detection,'' \emph{arXiv preprint arXiv:2110.08335}, 2021.

\bibitem[Liu et~al.(2021{\natexlab{b}})Liu, Gao, Zhang, Meng, and Lin]{liu2021investigating}
R.~Liu, J.~Gao, J.~Zhang, D.~Meng, and Z.~Lin, ``Investigating bi-level optimization for learning and vision from a unified perspective: A survey and beyond,'' \emph{IEEE Transactions on Pattern Analysis and Machine Intelligence}, vol.~44, no.~12, pp. 10\,045--10\,067, 2021.

\bibitem[Krizhevsky et~al.(2009)Krizhevsky, Hinton, et~al.]{krizhevsky2009learning}
A.~Krizhevsky, G.~Hinton \emph{et~al.}, ``Learning multiple layers of features from tiny images,'' \emph{Toronto, ON, Canada}, 2009.

\bibitem[Deng et~al.(2009)Deng, Dong, Socher, Li, Li, and Fei-Fei]{deng2009ImageNet}
J.~Deng, W.~Dong, R.~Socher, L.-J. Li, K.~Li, and L.~Fei-Fei, ``Imagenet: A large-scale hierarchical image database,'' in \emph{2009 IEEE conference on computer vision and pattern recognition}.\hskip 1em plus 0.5em minus 0.4em\relax Ieee, 2009, pp. 248--255.

\bibitem[He et~al.(2016)He, Zhang, Ren, and Sun]{he2016deep}
K.~He, X.~Zhang, S.~Ren, and J.~Sun, ``Deep residual learning for image recognition,'' in \emph{Proceedings of the IEEE conference on computer vision and pattern recognition}, 2016, pp. 770--778.

\bibitem[Touvron et~al.(2021)Touvron, Cord, Douze, Massa, Sablayrolles, and J{\'e}gou]{touvron2021training}
H.~Touvron, M.~Cord, M.~Douze, F.~Massa, A.~Sablayrolles, and H.~J{\'e}gou, ``Training data-efficient image transformers \& distillation through attention,'' in \emph{International conference on machine learning}.\hskip 1em plus 0.5em minus 0.4em\relax PMLR, 2021, pp. 10\,347--10\,357.

\bibitem[Liu et~al.(2020)Liu, Ma, Bailey, and Lu]{liu2020reflection}
Y.~Liu, X.~Ma, J.~Bailey, and F.~Lu, ``Reflection backdoor: A natural backdoor attack on deep neural networks,'' in \emph{Computer Vision--ECCV 2020: 16th European Conference, Glasgow, UK, August 23--28, 2020, Proceedings, Part X 16}.\hskip 1em plus 0.5em minus 0.4em\relax Springer, 2020, pp. 182--199.

\bibitem[Cheng et~al.(2021)Cheng, Liu, Ma, and Zhang]{cheng2021deep}
S.~Cheng, Y.~Liu, S.~Ma, and X.~Zhang, ``Deep feature space trojan attack of neural networks by controlled detoxification,'' in \emph{Proceedings of the AAAI Conference on Artificial Intelligence}, vol.~35, no.~2, 2021, pp. 1148--1156.

\bibitem[Tariq et~al.(2018)Tariq, Daniels, Schwartz, Washington, Kalantarian, and Wall]{tariq2018mobile}
Q.~Tariq, J.~Daniels, J.~N. Schwartz, P.~Washington, H.~Kalantarian, and D.~P. Wall, ``Mobile detection of autism through machine learning on home video: A development and prospective validation study,'' \emph{PLoS medicine}, vol.~15, no.~11, p. e1002705, 2018.

\bibitem[Grosse et~al.(2023)Grosse, Bieringer, Besold, Biggio, and Krombholz]{grosse2023machine}
K.~Grosse, L.~Bieringer, T.~R. Besold, B.~Biggio, and K.~Krombholz, ``Machine learning security in industry: A quantitative survey,'' \emph{IEEE Transactions on Information Forensics and Security}, vol.~18, pp. 1749--1762, 2023.

\bibitem[Zhang et~al.(2023)Zhang, Khanduri, Tsaknakis, Yao, Hong, and Liu]{zhang2023introduction}
Y.~Zhang, P.~Khanduri, I.~Tsaknakis, Y.~Yao, M.~Hong, and S.~Liu, ``An introduction to bi-level optimization: Foundations and applications in signal processing and machine learning,'' \emph{arXiv preprint arXiv:2308.00788}, 2023.

\bibitem[Neuhaus et~al.(2022)Neuhaus, Augustin, Boreiko, and Hein]{neuhaus2022spurious}
Y.~Neuhaus, M.~Augustin, V.~Boreiko, and M.~Hein, ``Spurious features everywhere--large-scale detection of harmful spurious features in imagenet,'' \emph{arXiv preprint arXiv:2212.04871}, 2022.

\bibitem[Sagawa et~al.(2020)Sagawa, Raghunathan, Koh, and Liang]{sagawa2020investigation}
S.~Sagawa, A.~Raghunathan, P.~W. Koh, and P.~Liang, ``An investigation of why overparameterization exacerbates spurious correlations,'' in \emph{International Conference on Machine Learning}.\hskip 1em plus 0.5em minus 0.4em\relax PMLR, 2020, pp. 8346--8356.

\bibitem[Sagawa et~al.(2019)Sagawa, Koh, Hashimoto, and Liang]{sagawa2019distributionally}
S.~Sagawa, P.~W. Koh, T.~B. Hashimoto, and P.~Liang, ``Distributionally robust neural networks for group shifts: On the importance of regularization for worst-case generalization,'' \emph{arXiv preprint arXiv:1911.08731}, 2019.

\bibitem[Bissoto et~al.(2020)Bissoto, Valle, and Avila]{bissoto2020debiasing}
A.~Bissoto, E.~Valle, and S.~Avila, ``Debiasing skin lesion datasets and models? not so fast,'' in \emph{Proceedings of the IEEE/CVF Conference on Computer Vision and Pattern Recognition Workshops}, 2020, pp. 740--741.

\bibitem[Moayeri et~al.(2022)Moayeri, Pope, Balaji, and Feizi]{moayeri2022comprehensive}
M.~Moayeri, P.~Pope, Y.~Balaji, and S.~Feizi, ``A comprehensive study of image classification model sensitivity to foregrounds, backgrounds, and visual attributes,'' in \emph{Proceedings of the IEEE/CVF Conference on Computer Vision and Pattern Recognition}, 2022, pp. 19\,087--19\,097.

\end{thebibliography}
}}

\appendix
\setcounter{section}{0}

\section*{Appendix}

\setcounter{section}{0}
\setcounter{figure}{0}
\makeatletter 
\renewcommand{\thefigure}{A\arabic{figure}}% Figure counter representation
\renewcommand{\theHfigure}{A\arabic{figure}}% Hyperref figure hyperlink hook
\renewcommand{\thetable}{A\arabic{table}}
\renewcommand{\theHtable}{A\arabic{table}}

\makeatother
\setcounter{table}{0}

\setcounter{mylemma}{0}
\renewcommand{\themylemma}{A\arabic{mylemma}}
\setcounter{equation}{0}
\renewcommand{\theequation}{A\arabic{equation}}

\section{Practical Conditions of various defense methods}
\label{app: idscore}

% \begin{table}
% \vspace*{-3mm}
% \centering
% \caption{\footnotesize{
% } 
% }
% \resizebox{\textwidth}{!}{
% \begin{tabular}{c|c}
% \toprule[1pt]
% \midrule
% Defense Method & Identification Scores & 
% \midrule
% ABL \citep{li2021anti} &  Training Loss  \\
% DBD \citep{huang2022backdoor} &  Symmetric Cross Entropy Loss  \\
% SD-FCT \citep{chen2022effective} &  FCT (feature consistency towards transformations) Metric \\
% STRIP \citep{gao2019strip} &  Entropy  \\
% SPC \citep{guo2023scaleup} &  SPC Loss \\
% CD \citep{huang2023distilling} &  Norm of optimized "minimal" mask \\
% Meta-SIFT \citep{zeng2022sift} & Weights assigned by a Meta Network \\
% \midrule
%  Scale-SIFT (Ours) & MSPC Loss \\
% \bottomrule[1pt]
% \end{tabular}
% }
% \vspace*{4mm}
% \end{table}%%%\\

\begin{table}[h]
\vspace*{-3mm}
\centering
\caption{\footnotesize{Practical conditions, different scoring and thresholding mechanisms used by different backdoor identification methods. $\gamma$ indicates the actual poisoning ratio. $\eps$ indicates an upper bound on the poisoning ratio. 
} 
}
\label{tab: idtable}
\resizebox{\textwidth}{!}{
\begin{tabular}{c|c|c|c|c}
\toprule[1pt]
\midrule
\multirow{3}{*}{Method} & \multicolumn{2}{c}{Realistic Conditions} & \multirow{3}{*}{Identification Scores} & \multirow{3}{*}{Thresholding  Method} \\
 & \textbf{(P1)} Free of & \textbf{(P2)} Free of & & \\
 & Clean Data &  Detection Threshold & & \\
\midrule
ABL \citep{li2021anti} & \cmark & \xmark & Training Loss & Heuristic\\
DBD \citep{huang2022backdoor}  & \cmark & \xmark & Symmetric Cross Entropy Loss & Heuristic\\
SD-FCT \citep{chen2022effective} & \cmark & \xmark & FCT Metric & Heuristic\\
STRIP \citep{gao2019strip}  & \xmark & \xmark  &  Entropy &  Estimated from Clean Data \\
AC \citep{chen2018detecting} & \cmark & \xmark & Silhoutte Scores of Activation Clusters  & Heuristic \\
\multirow{2}{*}{SS \citep{tran2018spectral}}  & \multirow{2}{*}{\cmark} & \multirow{2}{*}{\xmark} &  Inner Product between sample representation  & \multirow{2}{*}{1.5$\eps $}  \\
&&&and top singular vector of feature representation matrix&\\
SPECTRE \citep{hayase2021spectre} & \cmark & \xmark & QUEScore  & 1.5$\eps$ \\
SCAn \citep{tang2021demon} & \xmark & \cmark &  - & -\\
SPC \citep{guo2023scaleup}& \cmark & \xmark &  SPC Loss & Heuristic\\
CD \citep{huang2023distilling} & \xmark & \cmark &  Norm of optimized "minimal" mask & Estimated from Clean Data\\
Meta-SIFT \citep{zeng2022sift} & \cmark & \xmark & Weights assigned by a Meta Network & Manual based on $\gamma$\\
ASSET \citep{pan2023asset}  & \xmark & \cmark & Optimized Loss & Automated \\
\midrule
 Ours & \cmark & \cmark & MSPC Loss & \textbf{Automated (at 0 loss)}\\
\bottomrule[1pt]
\end{tabular}
}
\vspace*{4mm}
\end{table}%%%\\

In this section, we discuss the scores and the thresholding method used by relevant backdoor identification methods - which determines their satisfiability of the practical conditions defined in Section \ref{sec: intro}. 

Anti Backdoor Learning (ABL) \citep{li2021anti} consists of two steps - backdoor isolation and backdoor unlearning. In backdoor isolation, a local gradient ascent strategy is employed to amplify the difference in cross-entropy loss between the backdoor and benign samples. Based on a heuristic threshold on the training losses, a percentage of samples with low training losses are considered backdoor and further used for unlearning the backdoor samples. 

Backdoor Defense via Decoupling (DBD) \citep{huang2022backdoor} first learns a purified feature extractor using semi-supervised learning and then further learns a classification network on top of this extractor. This is done by the SCE (Symmetric Cross Entropy) loss, where the clean samples obtain a low loss on average than the backdoor samples. A heuristic threshold can be used to distinguish between potential backdoor and clean samples - e.g. $50 \%$ of the samples with the lowest loss are considered clean samples. 

The sample distinguishment (SD) module from \citep{chen2022effective} uses the FCT metric to distinguish between backdoor and clean samples, however these are used downstream for backdoor removal via unlearning or secure training via a modified contrastive learning. The FCT metric measures the distance between network features for a sample and its simple trasformation (like rotation etc.). \cite{chen2022effective} sets two thresholds that divides the FCT score into a poisoned set, a clean set and an uncertain set. 

Thus all of the above methods violate \textbf{P2} because of heuristic thresholds.

STRIP \citep{gao2019strip} finds the average entropy of a given sample over various perturbations (by superimposition with a base set of images). A low entropy (i.e. low randomness for change in prediction upon superposition) indicates a backdoor sample. However, to identify backdoor samples, a detection boundary needs to be set which is typically done by estimating the entropy distribution statistics of clean inputs, thus violating \textbf{P1} and \textbf{P2}.

% \SP{Activation Clustering
% In \citep{chen2018detecting}, Silhoutte scores of clusters from network activations }

Spectral Signature (SS) \citep{tran2018spectral} demonstrates the separability of poisoned and clean representations using correlations with the top singular vector of the representation matrix of samples. Thus, they compute an outlier score based on this to detect backdoor samples. Considering $\eps$ to be an upper bound on the number of backdoor samples, samples with the top $1.5\eps$ scores are considered backdoor. Since setting $\eps$ is heuristic (and requires an approximate knowledge of the poisoning ratio), this violates \textbf{P2}. 

SPECTRE \citep{hayase2021spectre} uses similar statistical properties of representations (whitened using robust estimation of mean and standard deviation of clean data) called QUantum Entropy Score (QUEscore). However, similar to \citep{tran2018spectral}, the top $1.5\eps$ scores are considered to be backdoor samples thus violating \textbf{P2}.

We note that SCAn \citep{tang2021demon}, does not use such identification scores - rather it relies on hypothesis testing to test whether representations of samples from a specific class follows a multivariate mixture distribution (denoting backdoor) or a single distribution (denoting benign). However, it uses clean data for the hypothesis testing and violates \textbf{P1}.

% A class of backdoor identification methods use separable latent representations learnt by a classifier to distinguish between backdoor and clean data points. Specifically, identification is achieved by \textit{separation} in silhoutte scores of clusters from network activations \citep{chen2018detecting}, or robust statistics \citep{tang2021demon,hayase2021spectre} or outlier scores computed using centered representation matrix and its top singular vector \citep{tran2018spectral}. While \cite{tang2021demon} assumes the presence of clean data, the other works use heuristic methods to threshold the mentioned score vectors. Moreover, these methods fundamentally rely on the \textit{latent separability assumption} , which was challenged through adaptive attacks in \cite{qi2023revisiting}, which mitigates this separation and subsequently, such identification methods. 

Scaled Prediction Consistency (SPC) \citep{guo2023scaleup} uses the scale-invariant backdoor signature (which we describe before). However, the detection threshold of the SPC loss (samples with loss above this threshold is considered backdoor) is either determined heuristically or through an additional set of clean data. In this paper, we consider the milder condition of unavailability of clean data - thus only using SPC violates \textbf{P2}. 

Cognitive Distillation (CD) \citep{huang2023distilling} uses the cognitive pattern signature. Cognitive pattern refers to the minimal distilled pattern (given by a mask) required for a particular sample to retain its original prediction from a trained model. The work showed that the $l_1$ norm of the CP of backdoor samples tend to be much smaller than that of clean samples. However, to set the threshold over the norms, the presence of a clean set is assumed, thus violating \textbf{P1}. 

In Meta-SIFT \citep{zeng2022sift}, a bi-level optimization algorithm is formulated that separates the clean and backdoor samples in the upper and lower level of their formulation using weights , similar to \eqref{eq: Eq1_v2} . However, the weights are given by a meta-network - based on which a fraction of clean samples are selected. Thus, in this scenario, to identify backdoor samples accurately, ideally one should know the poison ratio or they need to set a heuristic threshold (violating \textbf{P2}). 

\section{Importance of the Practical conditions}
\label{app: prac_imp}
In this section, we emphasize the importance of the practical conditions that we introduce in our paper. 

\textbf{Importance of P1 (Free of Clean Data)}:

We provide several examples to understand the importance of the clean data assumption:

\textbf{(1)} Firstly, we provide an example where it may be very hard / impossible to obtain clean data. We consider the case of prediction of the presence of Autism Spectrum Disorder (ASD) in children. Recent studies \citep{tariq2018mobile} collect home videos of children (submitted by parents) via crowdsourcing. Parents are required to take home videos of their child and upload them to a portal. Such crowdsourcing methods maybe highly susceptible to backdoor attacks.

In such cases, it may be very hard to collect clean data because that would require monitoring of the video capturing and uploading. This can be difficult or impossible because of various reasons (eg. change in behaviour of children, inavailability of consent to monitoring etc.)

\textbf{(2)} In some cases, organizations (which are clients to various ML companies) may be reluctant to provide / collect clean data for their application. For example, in a recent industry survey \citep{grosse2023machine}, companies included backdoor attacks as one of their security concerns. Such organizations tend to push the security responsibility upstream to service providers as mentioned in \citep{grosse2023machine} by one of the respondents - "We use Machine Learning as a Service and expect them to provide these robust algorithms and platforms”.

We believe many such scenarios associated with impossibility or reluctance of clean data collection may arise and are only limited to our imagination. This necessitates the development of algorithms without clean data assumption.

\textbf{Importance of P2 (Free of Detection Threshold)}:

We consider the case where this assumption is violated. Under such a condition, identification of backdoor datapoints requires the user to set some threshold on the scores assigned to each training data sample. Such a threshold depends on various factors like the poisoning ratio, the backdoor attack type and the dataset being used. For example, defense methods like SPECTRE \citep{hayase2021spectre} and Spectral Signature \citep{tran2018spectral} assume the knowledge of some upper bound of the poisoning ratio. On the other hand, CD \citep{huang2023distilling} suggests calculating the threshold from the mean and standard deviation of a subset of clean samples. We provide the various existing thresholding methods in \textbf{Table \ref{tab: idtable}} in the Appendix \ref{app: idscore}.

Thus, without some form of prior knowledge like the poisoning ratio (which is seldom the case), it is almost impossible for an user to determine such a threshold value. Thus, this indicates the importance of our assumption P2.
% This also shows that previous methods are infact not comparable.

\section{Details for the Bi-level Optimization Formulation}
\label{app: bilevel}

In this section, we will describe the strategies we adopt to develop the bi-level formulation \citep{zhang2023introduction} \eqref{eq: bilevel}. Recall the  data splitting problem of   \eqref{eq: Eq1_v2} :

\vspace*{-5mm}
{\small{
\begin{align}
\begin{array}{ll}
    \displaystyle  \min_{\mathbf w \in \{0,1\}^N}  & \sum_{i=1}^N (1-w_i) \ L(\mathbf x_i, \boldsymbol \mu^*(\mathbf{w}))  \\
     \ST  & \boldsymbol{\mu}^*(\mathbf{w}) = \argmax_{\boldsymbol \mu} \ \frac{1}{N}\sum_{i=1}^N w_i \ L(\mathbf x_i, \boldsymbol \mu)
\end{array}
\end{align}}}%

We consider masking variable $\mathbf{m}$ in MSPC as the optimization variable $\boldsymbol{\mu}$ in \eqref{eq: Eq1_v2}. In addition, the original SPC optimization is considered for an \textit{upper-level} problem that involves the lower-level solution $\boldsymbol{\mu}^*(\mathbf w)$, which is also a function of the upper-level variables $\mathbf w$.

Given the generic bi-level form in \eqref{eq: Eq1_v2},
both the upper and lower level optimizations work cohesively towards distinguishing between the backdoor samples and the clean ones. The lower level optimization tackles this challenge by amplifying the representation of backdoor samples within the loss. Conversely, the upper level optimization seeks to achieve the same goal by reducing the contribution of the poisoned samples within $L$. We note that this framework is different from bi-level optimization used for sifting out clean samples in  \cite{zeng2022sift}, where $L$ is simply the cross entropy loss and $\boldsymbol{\mu}$ represents the parameters of a neural network model.  We next need to integrate the proposed MSPC loss   with the bi-level framework \eqref{eq: Eq1_v2}.

However, the MSPC loss is non-differentiable because of the occurrence of the function $\phi(x)$ in \eqref{eq: MSPC}. We will now describe the strategies we adopt in each optimization level, to mitigate such issues, leading to our proposed bi-level solution.

First, we note that the basis of scaled prediction consistency is that for poisoned samples, the predicted class of a model for an input $\mathbf{x}$ remains consistent even when multiplied with scales, \textit{i.e.}, $\argmax\mathcal{F}_{\boldsymbol \theta}(\cdot)$ remains the same. Thus, for a sample $\mathbf{x}_i$ and for a scale $n \in \mathcal{S}$, maximizing the SPC loss would result in maximizing a non-differentiable loss $\mathbbm{1}(\displaystyle \argmax\mathcal{F}_{\boldsymbol \theta}(\mathbf{x}) = \displaystyle \argmax\mathcal{F}_{\boldsymbol \theta}(n\cdot\mathbf{x}))$ for SPC or maximizing the non-differentiable $\phi(\displaystyle \argmax\mathcal{F}_{\boldsymbol \theta}(\mathbf{x}) - \displaystyle \argmax\mathcal{F}_{\boldsymbol \theta}(n\cdot\mathbf{x}))$ for the MSPC loss. In such a scenario, we argue that the probability distribution that we get from the model for poisoned inputs across scales would also remain similar, \textit{i.e.}, the distance between $\mathcal{F}_{\boldsymbol \theta}(\mathbf{x})$ and $\mathcal{F}_{\boldsymbol \theta}(n\cdot \mathbf{x})$ would remain small when $\mathbf{x} \in \mathcal{D}_b$.  Based on that, we can simply replace the non-differentiable loss maximization with the minimization of the Kullback-Leibler (KL) divergence. Specifically, this involves minimizing the divergence between the probability distribution output for a sample $\mathbf{x}_i$ and the output for the same sample when masked with $\mathbf{m}$ and subsequently scaled with $n$. This is in line with the proposed use of masks $\mathbf{m}$. In mathematical terms, this is denoted as $\min_{\mathbf{m}} D_{KL} (\mathcal{F}_{\boldsymbol \theta}(\mathbf{x}_i) || \mathcal{F}_{\boldsymbol \theta}(n \cdot\mathbf{x}_i^\mathbf{m}))$. 

Leveraging the advantages of the proposed linear shift by $\tau$, we can customize the \textbf{lower-level optimization} of \eqref{eq: Eq1_v2}  to
% \vspace*{0.1mm}

\vspace*{-3mm}
{\small{
\begin{align}
 \mathbf{m}^*(\mathbf w) = \argmin_{\mathbf{m}} \ \frac{1}{N}\sum_{i=1}^N  \Big[ w_i\ \frac{1}{|S|}\sum_{n \in S} D_{KL} (\mathcal{F}_{\boldsymbol \theta}(\mathbf{x}_i) || \mathcal{F}_{\boldsymbol \theta}(n \cdot\mathbf{x}_i^{m}))\Big] + \lambda \|\mathbf{m}\|_1 , 
    \label{eq: Eq2_app}
\end{align}}}%
where recall that $\mathbf{x}_i^{m} = (\mathbf{x}_i - \tau) \odot \mathbf{m}$.
The above addresses the challenge of optimizing the mask variable in a non-differentiable loss function by converting the lower-level optimization to a differentiable one,   achieved through utilizing KL divergence. 

As mentioned in Section \ref{sec: PropMethod}, we simply use the MSPC loss for the \textbf{upper-level optimization}, thus achieving \eqref{eq: bilevel}. 

We note that the upper level is intricately linked with the lower level optimization, rendering the problem more complex. In practical terms, we managed to simplify this by treating the $\mathbf{m}^*(\mathbf{w})$ term in the upper optimization as a constant value, considering only the $\mathbf w$ variable outside the MSPC loss. In this way, we can still use the MSPC loss in the upper-level optimization. 

To solve the problem, we initialize the mask with the value of 1 for all pixels. We note that minimizing the upper level optimization is trivial in our case: we set $w_i = 1$ if $\ell_\mathrm{MSPC} (\mathbf{x}_i, \mathbf{m} ,\tau)$ is positive and $w_i = 0$ if $\ell_\mathrm{MSPC} (\mathbf{x}_i, \mathbf{m} ,\tau)$ is negative. We choose an odd number of scales for our experiments so that the value of the MSPC loss is never 0. During this upper level optimization, we consider the mask $\mathbf{m(w)}$ to be fixed. Solving the upper level optimization yields us a $\mathbf{w}$, which we use to solve the lower level optimization using gradient descent. We can perform this process iteratively to find the final MSPC loss vector where the $i$th value of this vector is given by $\ell_\mathrm{MSPC} (\mathbf{x}_i, \mathbf{m}^*(\mathbf{w}^*),  \tau)$. This approach of alternating optimization, although a compromise, yielded satisfactory results (as demonstrated in Section \ref{sec: expres}) and significantly reduced the complexity of our optimization process. 

\section{Understanding Scenario 3 in Section\,\ref{sec: PropMethod}}
\label{app: scen3}

We revisit the phenomenon of the  Scenario 3 (\textbf{Figure \ref{fig: spclimit}}-CIFAR10 row 5) that we described in Section \ref{sec: PropMethod}. We would like to provide a more detailed explanation of why the observed phenomenon reveals a spurious correlation. 

Neural networks often rely on {spurious features} for their prediction \citep{neuhaus2022spurious}. The problem of spurious correlation can come from strong association between labels and backgrounds in classification \citep{sagawa2020investigation} (\textit{e.g.}, Waterbird or CelebA dataset) \citep{sagawa2019distributionally}, or by relying on artifacts \citep{bissoto2020debiasing} or simply by background reliance \citep{moayeri2022comprehensive}. We discover an intriguing phenomenon similar to the effect of spurious correlation: Certain spurious image patterns can be found simply when scaling the images by scalars. The predictions of images remain intact because of their reliance on such spurious features, as shown in \textbf{Figure\,\ref{fig: spclimit}}-column CIFAR-10 row 5. We emphasize that this is not a systematic detection of spurious features, rather an observation which explains high SPC values for benign samples.

\begin{figure}[h]
\centerline{
\begin{tabular}{c}
    \hspace*{-2mm}  \includegraphics[width=\textwidth,height=!]{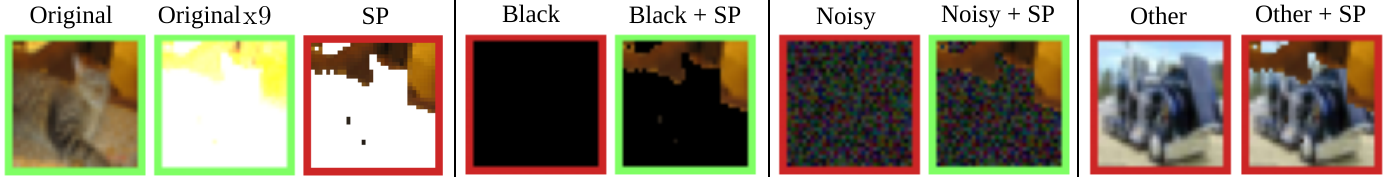} 
\end{tabular}}
%\vspace*{-3.3mm}
\caption{\footnotesize{Figure indicates the presence of Spurious Correlation. `$\mathbf{x} +$ SP' indicates an image formed by stamping the Spurious Pattern (SP) into $\mathbf{x}$. 
% \SL{Here SP is obtained by multiplying the original `cat' image with the scaling factor $9$.} 
Here SP is obtained by masking out the pixels of the original `cat' image which stay (\textit{i.e.} pixel values < 1 across all 3 channels), when multiplied with the scaling factor $9$. We display SP with a white background. `Black' denotes a completely black image (with zero pixel values). `Noisy' denotes an image drawn from a uniform distribution between 0 and 0.3. `Other' indicates another image from the dataset CIFAR-10. The \setlength{\fboxsep}{0pt}\colorbox{lightgreen}{green box} indicates that the predicted label is the same as `cat' and the \setlength{\fboxsep}{0pt}\colorbox{lightred}{red box} indicates that the predicted label is different from `cat'.
}}
  \label{fig: spuriouspattern}
 \vspace*{-2mm}
\end{figure}%%%

In \textbf{Figure\,\ref{fig: spclimit}}-row 5, we observed that the main features of the original class (\textit{i.e.} the cat) vanished when multiplied with higher scalar values. However, the model continued predicting the image as a cat.  
%(\textit{i.e.} class 3). 
We notice that a yellow patch in that Figure remained. This naturally leads to the question of \textit{whether that remaining patch causes the model to classify the image as cat}.
To investigate this, we track the model's prediction when this patch is stamped onto various backgrounds. As seen in \textbf{Figure\,\ref{fig: spuriouspattern}}, when this patch is pasted in a black or slightly noisy image, the model still predicts it as the class of cat, in spite of the fact that there exist no predictive features of a cat. This indicates the presence of a spurious correlation. However, as seen in \textbf{Figure\,\ref{fig: spuriouspattern}}-column 4, the spurious correlation is not strong enough to affect the model prediction in the presence of other objects.

\section{Training Details.}
\label{app: hpconfig}

\begin{table*}[h]
% \vspace{-12mm}
\centering
  \caption{\footnotesize{Details for training models with different datasets used in this paper}}
\label{tab: tprfpr0.5}
  \resizebox{\textwidth}{!}{
  \begin{tabular}{cccccccc}
    \toprule
    \multirow{2}{*}{Dataset} & \multirow{2}{*}{Epochs} & Initial  & Learning Rate  & Learning Rate  & Learning Rate  & \multirow{2}{*}{Momentum} & Weight \\
      &  & Learning Rate & Scheduler  & Decay Factor & Decay Epoch &  &  Decay\\
    \midrule
    \midrule
    CIFAR-10 & 182 & 0.1 & MultiStep LR & 0.1 & 91,136 & 0.9 & 5e-4\\
    TinyImageNet & 100 & 0.1 & MultiStep LR & 0.1 & 40,80 & 0.9 & 5e-4\\
    ImageNet200 & 60 & 0.5 & Cyclic LR & - & - & 0.9 & 5e-4\\
    \bottomrule
  \end{tabular}
  }    
\end{table*}
We notice that $\tau = 0.1$ is sufficient to ensure the success of our method, where $\tau$ is the small linear shift of a sample $\mathbf{x}$  as proposed in \eqref{eq: MSPC}. As mentioned in Section \ref{sec: PropMethod}, We choose a series of scaling factors $\{2,3,...,12\}$ following \citep{guo2023scaleup} in our experiments. For CIFAR-10 experiments, we solve our bi-level optimization problem using 10 epochs of minibatch SGD with a batch size of 1000 and a learning rate of 0.1 for the lower level, with a momentum of 0.9 and a weight decay ratio of $5\times10^{-4}$. We perform the optimization for a total of 4 outer epochs. For TinyImageNet and ImageNet200 experiments, we perform only 5 epochs for the lower level. All experiments are conducted using 4 NVIDIA GeForce RTX 3090 GPUs with Pytorch implementation. We used FFCV (see \href{https://ffcv.io/}{https://ffcv.io/}) to speed up our training process.

\section{Backdoor Attack Setting Details}
\label{app: attacksetting}

We consider 6 widely used backdoor attacks for our experiments in this paper. We describe the detailed settings of each of those attacks as follows.

\begin{itemize}[leftmargin=*]
    \item Badnet \citep{gu2017badnets} : We consider a $5 \times 5$ RGB trigger at the bottom right corner of our images.  
    \item Blend \citep{chen2017targeted} : As reported in the literature \citep{chen2017targeted}, we use random patterns as triggers and blend them in the image. Specifically, for an image $\mathbf{x}$ and the trigger pattern $\mathbf t$, the backdoor sample $\mathbf{x}^\prime = 0.8\mathbf{x} + 0.2\mathbf{t}$. 
    \item LabelConsistent \citep{turner2019label} : This attack is performed by adding adversarial perturbations, generated using projected gradient descent bounded by $\ell_{\infty}$ norm, to images from the target class.
    % \SL{[Check the following revised sentence. I did not fully understand the original statement.]}
    We directly follow the above generation process detailed in \citep{huang2023distilling}. 
    % \SL{[why do you still add trigger at the presence of using perturbations?]} \SP{The adv perturbations are responsible for making it harder for the model to use the generalizing features of the target class and the trigger establishes the correlation between trigger and target class.}
    Moreover, the backdoored images are also stamped with $3 \times 3$ trigger (pixel values 0 or 255) at the bottom right corner. 
    \item TUAP \citep{zhao2020clean}: This attack involves a universal adversarial perturbation of images from the target class. This was performed using a 10-step projected gradient descent $\ell_{\infty}$ attack on a clean trained ResNet-18 model with maximum perturbation of $\epsilon = 8/255$ and a learning rate of $2/255$. Additionally, a Badnet-like trigger was attached to those images for the backdoor attack.
    \item Trojan \citep{liu2018trojaning} : We use a trigger given by \cite{li2021neural}, which was constructed by reverse-engineering the last fully-connected layer of the network as proposed in the original paper \citep{liu2018trojaning}.
    \item Wanet \citep{nguyen2021wanet} : We follow the framework of \citep{nguyen2021wanet} to generate a grid that is used to warp images to form the corresponding backdoor samples, apart for a few changes to ensure successful poisoning in our framework. Specifically, we start with an uniform grid of size $32 \times 32$ (similar to \cite{huang2023distilling}) with random values between $-1$ and $+1$. We choose the warping strength \citep{nguyen2021wanet} to be 0.5, which is consistent with the setting in the original paper. This grid acts as the flow field, used to warp images. Similar to \cite{huang2023distilling}, we also do not manipulate the training objective, thus staying consistent with our framework.
    \item DFST \citep{cheng2021deep} : This involves a style transfer of the input image using a cycleGAN. The style transfer is done using a 'sunrise weather' style used by \cite{cheng2021deep}. We use the poisoned data provided by \cite{huang2023distilling}.
    \item AdapBlend \citep{qi2023revisiting} : The Adaptive Blend attack is performed similar to the Blend attack using the 'hellokity' blending pattern with modifications to reduce the latent separability between clean and backdoor samples. Specifically, the trigger is partitioned into 16 pieces - of which $50 \%$ are used at train time and all are used at test time. Secondly, among the samples that are blended with the trigger, the labels of $50 \%$ of them are replaced with the target label , while the others preserve the actual label (conservatism ratio = 0.5). Finally, an extremely low poisoning ratio of 0.003 is considered for this attack.
    \end{itemize}

% \begin{table*}[htb]
%  \vspace*{-2mm}
%   \centering
%   \caption{\footnotesize{Average standard accuracy (SA) and attack success rate (ASR) for different attacks. Standard deviation is given in parenthesis and the values of 
%   mean and standard deviation are calculated over 3 runs. Here $\gamma$ refers to the poisoning ratio.}}
% \label{tab: accasr}
%   \resizebox{0.7\textwidth}{!}{
%   \begin{tabular}{ccccc}
%     \toprule
%      \multirow{2}{*}{Backdoor Attack}&\multicolumn{2}{c}{$\gamma = 0.1$} &\multicolumn{2}{c}{$\gamma = 0.05$}\\
%      \cmidrule(lr){2-3} \cmidrule(lr){4-5} 
%      &  SA (\%)       &       ASR (\%)     &         SA (\%)       &      ASR (\%)  \\
%     \midrule
%     \midrule
%     Badnet & 88.36 (0.12) & 100.00 (0.00) &88.77 (0.15) & 100.00 (0.00)\\ 
%  Blend & 88.29 (0.32) & 100.00 (0.00) &88.75 (0.19) & 100.00 (0.00)\\ 
%  LabelConsistent & 89.36 (0.42) & 99.31 (0.33) &89.14 (0.21) & 96.66 (1.31)\\ 
%  TUAP & 89.18 (0.39) & 99.46 (0.04) &89.08 (0.23) & 98.43 (0.93)\\ 
%  Trojan & 88.62 (0.09) & 100.00 (0.00) &88.75 (0.25) & 99.99 (0.01)\\ 
%  Wanet & 88.20 (0.20) & 98.53 (0.18) &88.90 (0.15) & 98.89 (0.15)\\ 

%     \bottomrule
%   \end{tabular}
%   }
% \end{table*}

%To ensure that the  poisoning effect did not vanish, we train the models without any data augmentation. 
We use the target label of class 1 for the label-corrupted attacks in our experiments. All models trained using the backdoor datasets can achieve an acceptable clean accuracy and can be successfully attacked. 
% as shown in Table \ref{tab: tprfpr}. 

\section{Baseline Defense Configurations.}
\label{app: defensedetail}

We describe the detailed settings of the baselines considered in this paper as follows: 

\begin{itemize}
    \item  SPC \citep{guo2023scaleup} : We consider the scaling factors $\{2,3,...,12\}$ to compute the SPC loss and use a threshold of 0.5 to find the TPR / FPR values
    \item ABL \citep{li2021anti} : We consider the flooding loss (to avoid overfitting as described in  \citep{li2021anti}) in the gradient ascent step with a flooding parameter of 0.5. Then we consider 10 $\%$ of samples with the lowest losses over 10 epochs as backdoor samples.
    \item SD-FCT \citep{chen2022effective} : The Feature Consistency metric is computed using a transformation set consisting of random 180 degrees rotation and random affine transformations. 
    \item STRIP \citep{gao2019strip} : To calculate the entropy, we took 100 randomly selected clean images from the validation set and used them to blend in the samples with equal weightage to both. 
\end{itemize}

\newpage

% \section{ROC plots for $\gamma = 0.05$.}
\section{Additional Results}
\label{app: addres}

\begin{figure}[!h]
\vspace*{-5mm}
%\vspace*{-6mm}
\centerline{
\begin{tabular}{cccc}
    \hspace*{-2mm}  \includegraphics[width=0.33\textwidth,height=!]{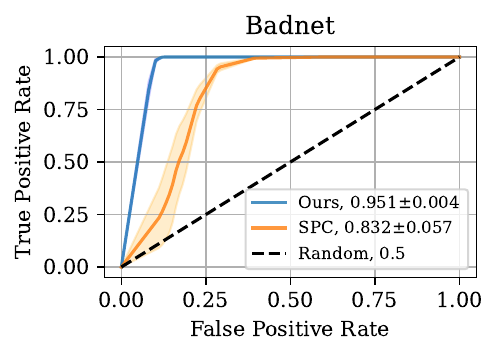} &
    \hspace*{-5mm} \includegraphics[width=0.33\textwidth,height=!]{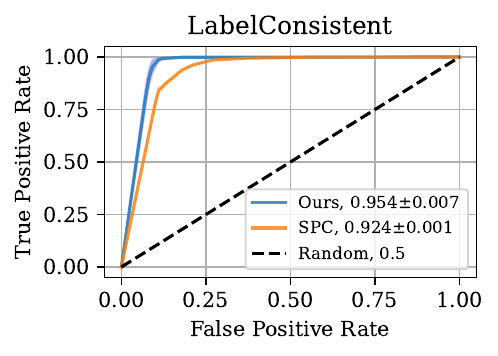} &
    \hspace*{-5mm} \includegraphics[width=0.33\textwidth,height=!]{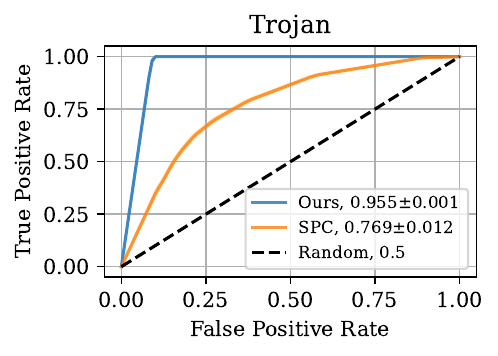} 
\end{tabular}}
%\vspace*{-3.3mm}
\caption{\footnotesize{Additional ROC plots for our method with reference SPC \citep{guo2023scaleup} using CIFAR-10. Error bars indicate half a standard deviation over 3 runs. The poisoning ratio is 0.1.
}}
  \label{fig: auroc_app}
 \vspace*{-2mm}
\end{figure}%%%

\begin{figure}[!h]
\vspace*{-5mm}
%\vspace*{-6mm}
\centerline{
\begin{tabular}{cccc}
    \hspace*{-2mm}  \includegraphics[width=0.33\textwidth,height=!]{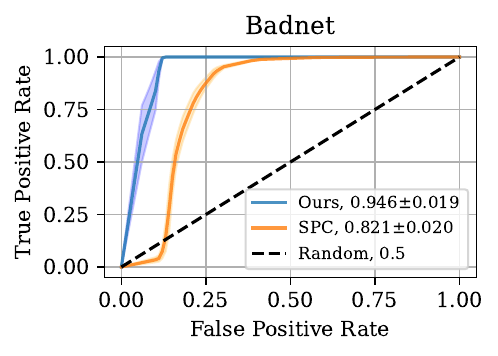} &
    \hspace*{-5mm} \includegraphics[width=0.33\textwidth,height=!]{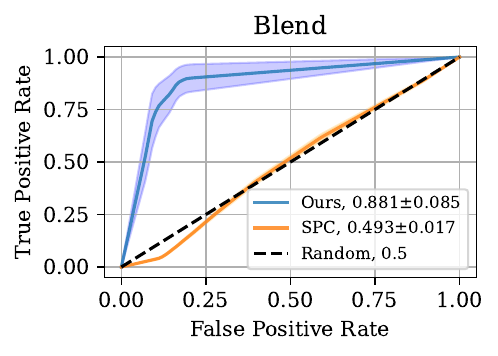} &
    \hspace*{-5mm} \includegraphics[width=0.33\textwidth,height=!]{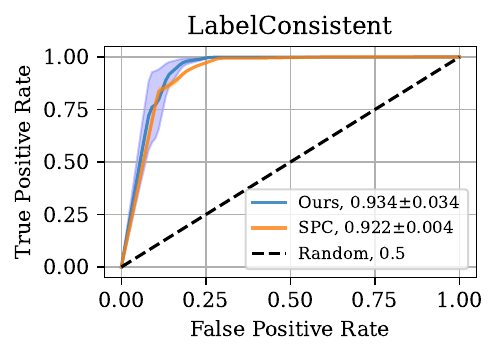} \\
    \hspace*{-2mm}  \includegraphics[width=0.33\textwidth,height=!]{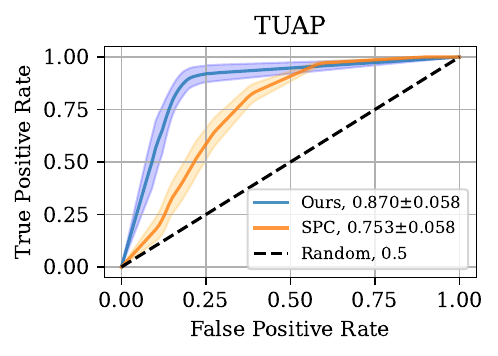} &
    \hspace*{-5mm} \includegraphics[width=0.33\textwidth,height=!]{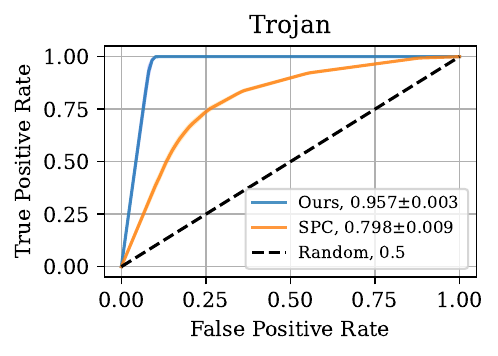} &
    \hspace*{-5mm} \includegraphics[width=0.33\textwidth,height=!]{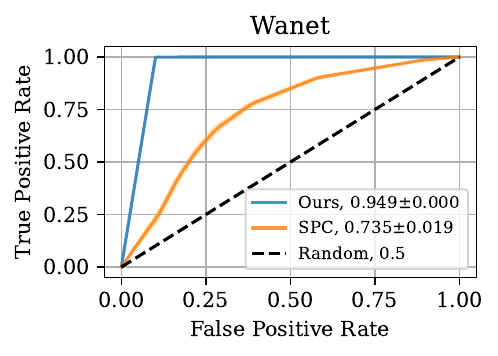}
\end{tabular}}
%\vspace*{-3.3mm}
\caption{\footnotesize{ROC plots for our method with reference SPC \citep{guo2023scaleup} using CIFAR-10. Error bars indicate half a standard deviation over 3 runs. The poisoning ratio is 0.05.
}}
  \label{fig: auroc_0.05}
 \vspace*{-2mm}
\end{figure}%%%

\begin{table*}[h]
% \vspace{-12mm}
\centering
  \caption{\footnotesize{Average TPR/FPR values for our method and for reference SPC \citep{guo2023scaleup}. Our algorithm has an \textbf{automated threshold of 0}. The threshold considered for SPC is 0.5. Results  are presented in the format of `TPR/FPR', where  the standard deviation is given in parenthesis, and the values of 
  mean and standard deviation are calculated over 3 runs. Poisoning ratio is 0.05}}
\label{tab: tprfpr0.5}
  \resizebox{0.7\textwidth}{!}{
  \begin{tabular}{ccc}
    \toprule
     Attack &  SPC       &        Ours    \\
    \rowcolor{LightCyan}
    \midrule
\multicolumn{3}{c}{\textbf{CIFAR-10}} \\
    \midrule
 Badnet           & 0.680 (0.093) / 0.186 (0.008)&1.000 (0.000) / 0.166 (0.007)\\ 
  Blend             & 0.128 (0.013) / 0.186 (0.005)&0.831 (0.184) / 0.132 (0.022)\\ 
  LabelConsistent   & 0.931 (0.010) / 0.190 (0.005)&0.984 (0.023) / 0.137 (0.049)\\ 
  TUAP              & 0.409 (0.163) / 0.182 (0.003)&0.859 (0.138) / 0.163 (0.044)\\
  Trojan           & 0.608 (0.028) / 0.172 (0.003)&0.999 (0.001) / 0.100 (0.001)\\ 
  Wanet           &0.465 (0.028) / 0.185 (0.006)&0.999 (0.001) / 0.101 (0.001)\\

    \bottomrule
  \end{tabular}
  }    
\end{table*}

We present the  ROC plots \textbf{Figure\,\ref{fig: auroc_app}} for additional attacks with the poisoning ratio of $\gamma = 0.1$ (not included in \textbf{Figure \ref{fig: auroc}}. Additionally, we also provide additional results  for poisoning ratio of $\gamma = 0.05$ in \textbf{Figure\,\ref{fig: auroc_0.05}}, \textbf{Table \ref{tab: tprfpr0.5}}. Similar to our observations in Section \ref{sec: expres}, our method reaches a near 1 TPR at a much lower FPR value.

\newpage

\section{ Results with ViT}
\label{app: vit}

We provide additional results using the Badnet attack on various datasets with a vision transformer in \textbf{Table\,\ref{tab: aurocvit}}. We consider the ViT-Ti/16 \citep{touvron2021training} pretrained model and finetune it for 50 epochs using Adam optimizer (learning rate = $1e-4$ and weight decay = $5e-4$) and a cosine annealing schedule. 

\begin{table}[h]
% \vspace{-5mm}
\centering
  \caption{\footnotesize{Mean AUROC values for our method and for reference SPC \citep{guo2023scaleup}. on CIFAR10, TinyImageNet, and ImageNet200 for the Badnet attack. The best result in each setting is marked in \textbf{bold}. All attacks have a poisoning ratio equal to 10\%}}
\label{tab: aurocvit}
  \resizebox{0.5\textwidth}{!}{
  \begin{tabular}{ccc}
    \toprule
     Attack &  SPC       &        Ours    \\
    % \rowcolor{LightCyan}
    \midrule
% \multicolumn{3}{c}{\textbf{CIFAR-10}} \\
    \midrule
  CIFAR-10   & 0.9045 (0.0160) & 0.9798 (0.0025)\\
  TinyImageNet   &  0.9037 (0.0034) & 0.9986 (0.0008) \\ 
  ImageNet200   & 0.9173 (0.0371) & 0.9982 (0.0006)\\
    \bottomrule
  \end{tabular}
  }    
  \vspace{-5mm}
\end{table}

\section{Ablation Study}
\label{app: ablation}

\begin{figure}
    \centering
    \includegraphics[width=0.5\textwidth,height=!]{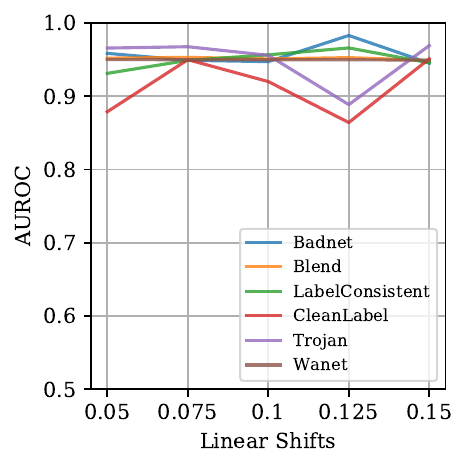}
    \caption{Variation of AUROC for different values of $\tau$ for different backdoor attacks.}
    \label{fig:ablation}
\end{figure}

We have included an ablation study for the hyperparameter $\tau$ , which is the linear shift as shown in \textbf{Figure \ref{fig:ablation}} of the attached document. We perform this experiment on CIFAR-10 using Resnet-18. We observe that our method achieves an AUROC>0.9 for most linear shifts across a diverse set of attacks.

\section{Additional Visualizations}
\label{app: addvis}

\begin{figure}[!h]
\vspace*{-0mm}
% \vspace*{-0mm}
\centerline{
% \hspace*{-9mm} 
\includegraphics[width=0.86\textwidth,height=!]{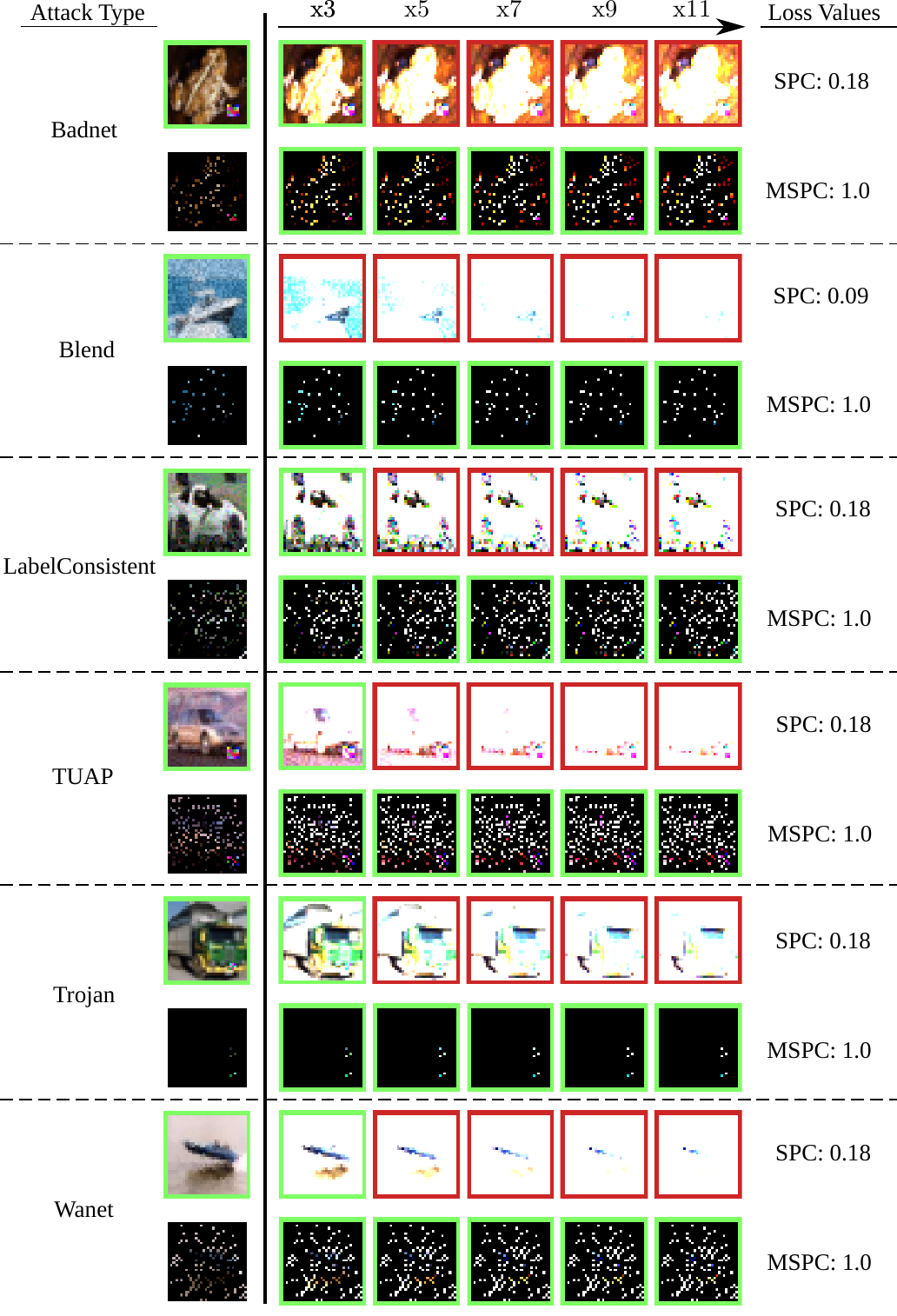}}
%\vspace*{-3.3mm}
\caption{\footnotesize{Additional visualizations of \textbf{backdoored images} with low SPC loss but high MSPC loss, across different types of backdoor attacks-poisoned training sets.  We present examples of the original image and the masked image across scales. The \textcolor{green}{green box} indicates that the predicted label is the target label and the \textcolor{red}{red box} indicates that the predicted label is different from  the target label. We set a threshold of 0.08 for masks for better visualisation. }}
  \label{fig: extravis1}
 \vspace*{-2mm}
\end{figure}

In this section, we present additional figures across a variety of backdoor attacks demonstrating the benefits of the masks (from our bi-level formulation) and algorithm as mentioned in Section \ref{sec: expres}. As shown in \textbf{Figure\,\ref{fig: extravis1}}, our optimized masks can in fact, focus on the triggers. For non-patch based triggers like Blend (\textbf{Figure\,\ref{fig: extravis1}}-row 2) and Wanet (\textbf{Figure\,\ref{fig: extravis1}}-row 6), we hypothesize that our algorithm finds the `effective part' of the trigger as explained in our formulation of the MSPC loss in Section \ref{sec: PropMethod}.

%%%

\begin{figure}[!h]
% \vspace*{-5mm}
\vspace*{-0mm}
\centerline{
% \hspace*{-9mm} 
\includegraphics[width=0.9\textwidth,height=!]{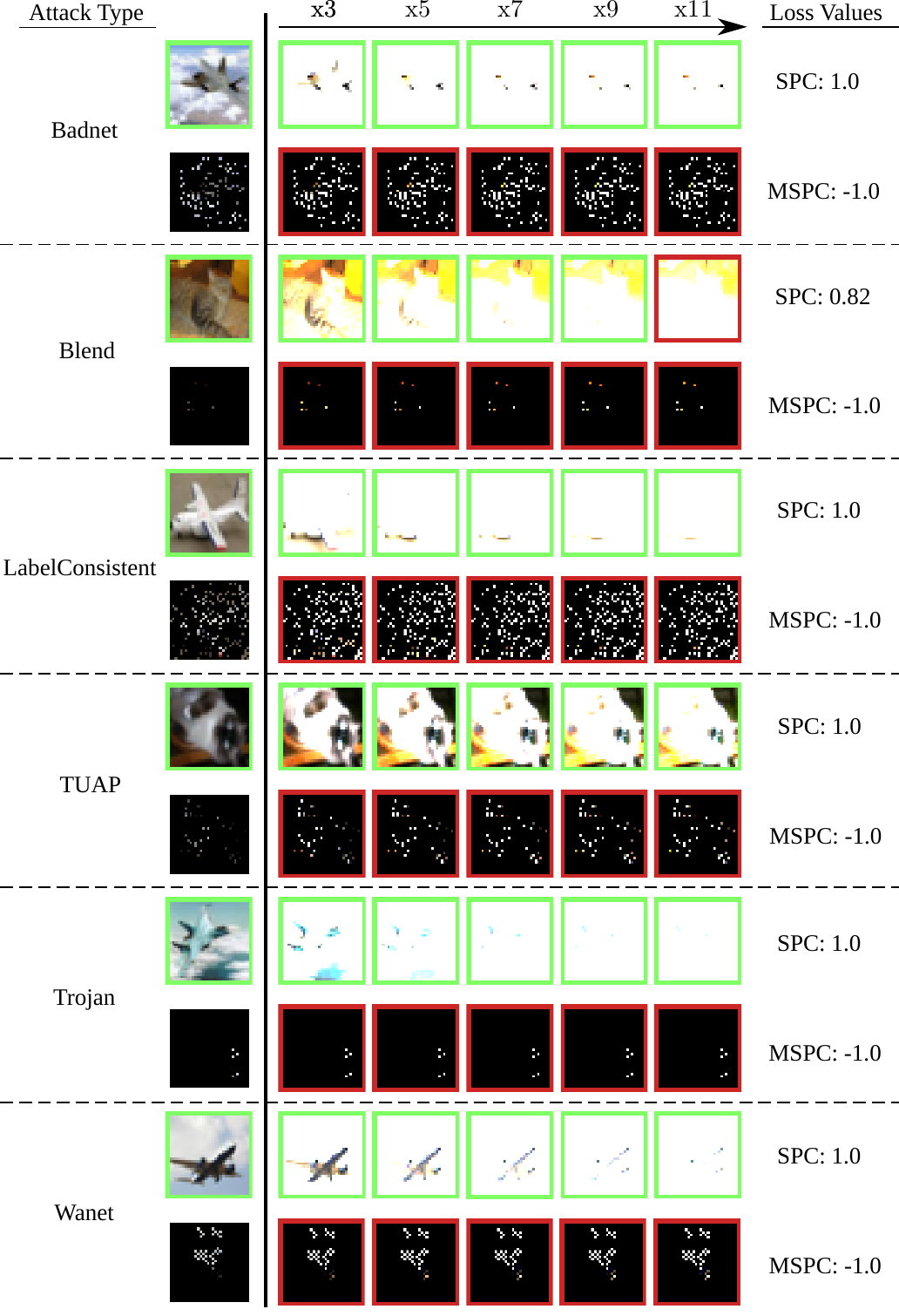}}
%\vspace*{-3.3mm}
\caption{\footnotesize{Additional visualizations of
\textbf{clean images} with high SPC loss but low MSPC loss, across different types of backdoor attacks-poisoned training sets. We use different masks learned after solving our proposed bi-level formulation of \eqref{eq: bilevel} as indicated by the attack names.  We present examples of the original image and the masked image across scales. The \textcolor{green}{green box} indicates that the predicted label is the same as the true class label of the image and the \textcolor{red}{red box} indicates that the predicted label is different from  the true class label of the image. \textit{ We note} that a high SPC indicates backdoor images,  and the clean images are predicted as backdoored samples because the prediction remains consistent across scales as indicated by the \textcolor{green}{green boxes}. Our proposed MSPC formulation prevents such a phenomenon through the use of our learned masks as indicated by the \textcolor{red}{red boxes}. We set a threshold of 0.08 (0.001 for Wanet) for masks for better visualisation.}}
  \label{fig: extravis2}
 \vspace*{-2mm}
\end{figure}%%%

 We also uncovered various cases of clean images with a high SPC value in  Section \ref{sec: PropMethod}. In \textbf{Figure\,\ref{fig: extravis2}}, we present various such examples where our formulation is successful. \textbf{Figure\,\ref{fig: extravis2}}-row 1 shows that our masks can avoid the white pixels caused due to the airplane vanishing (Section \ref{sec: PropMethod}-Scenario 2). As explained in Section \ref{sec: PropMethod}, our masks can also avoid the spurious correlation (\textbf{Figure\,\ref{fig: extravis2}}-row 2). We note that the mask in this case and in \textbf{Figure\,\ref{fig: mspcgood}} - CIFAR10,row 2 are different because we choose masks learnt from different runs. Lastly, our masks can also avoid the generalizing features that stays despite high scales (Section \ref{sec: PropMethod}-Scenario 1) as seen in \textbf{Figure\,\ref{fig: extravis2}}-row 4. Thus, our proposed formulation is successful in improved performance and automatic identification of backdoor datapoints, thus satisfying \textbf{P1} and \textbf{P2} (Section \ref{sec: intro}).

\end{document}